\documentclass[11pt]{article}

\usepackage{arabtex}

\usepackage{coling2018}

\usepackage{url}
\usepackage{float}
\usepackage{graphicx,subcaption}
\usepackage{smartdiagram}
\usepackage{pgf-pie}
\usepackage{caption} 

\usepackage[stable]{footmisc}

\usepackage{cp1256,utf8}
\setcode{utf8}

\date{}

\begin{document}

\title{Making Old Kurdish Publications Processable by 	Augmenting Available Optical Character Recognition	Engines}


\author{
	\begin{tabular}[t]{c}
		Blnd Yaseen and Hossein Hassani\\
		\textnormal{University of Kurdistan Hewl\^er}\\
		\textnormal{Kurdistan Region - Iraq}\\
		{\tt {\{blnd.yaseen, hosseinh}\}@ukh.edu.krd}
	\end{tabular}
}

\maketitle

\begin{abstract}
Kurdish libraries have many historical publications that were printed back in the early days when printing devices were brought to Kurdistan. Having a good Optical Character Recognition (OCR) to help process these publications and contribute to the Kurdish language’s resources which is crucial as Kurdish is considered a low-resource language. Current OCR systems are unable to extract text from historical documents as they have many issues, including being damaged, very fragile, having many marks left on them, and often written in non-standard fonts and more. This is a massive obstacle in processing these documents as currently processing them requires manual typing which is very time-consuming. In this study, we adopt an open-source OCR framework by Google, Tesseract version 5.0, that has been used to extract text for various languages. Currently, there is no public dataset, and we developed our own by collecting historical documents from Zheen Center for Documentation and Research, which were printed before 1950 and resulted in a dataset of 1233 images of lines with transcription of each. Then we used the Arabic model as our base model and trained the model using the dataset. We used different methods to evaluate our model, Tesseract’s built-in evaluator lstmeval indicated a Character Error Rate (CER) of 0.755\%. Additionally, Ocreval demonstrated an average character accuracy of 84.02\%. Finally, we developed a web application to provide an easy- to-use interface for end-users, allowing them to interact with the model by inputting an image of a page and extracting the text. Having an extensive dataset is crucial to develop OCR systems with reasonable accuracy, as currently, no public datasets are available for historical Kurdish documents; this posed a significant challenge in our work. Additionally, the unaligned spaces between characters and words proved another challenge with our work. 
\end{abstract}

\section{Introduction}
\label{sec:intro}

\begin{figure}[ht]
	\centering
	\frame{\includegraphics[scale=0.75]{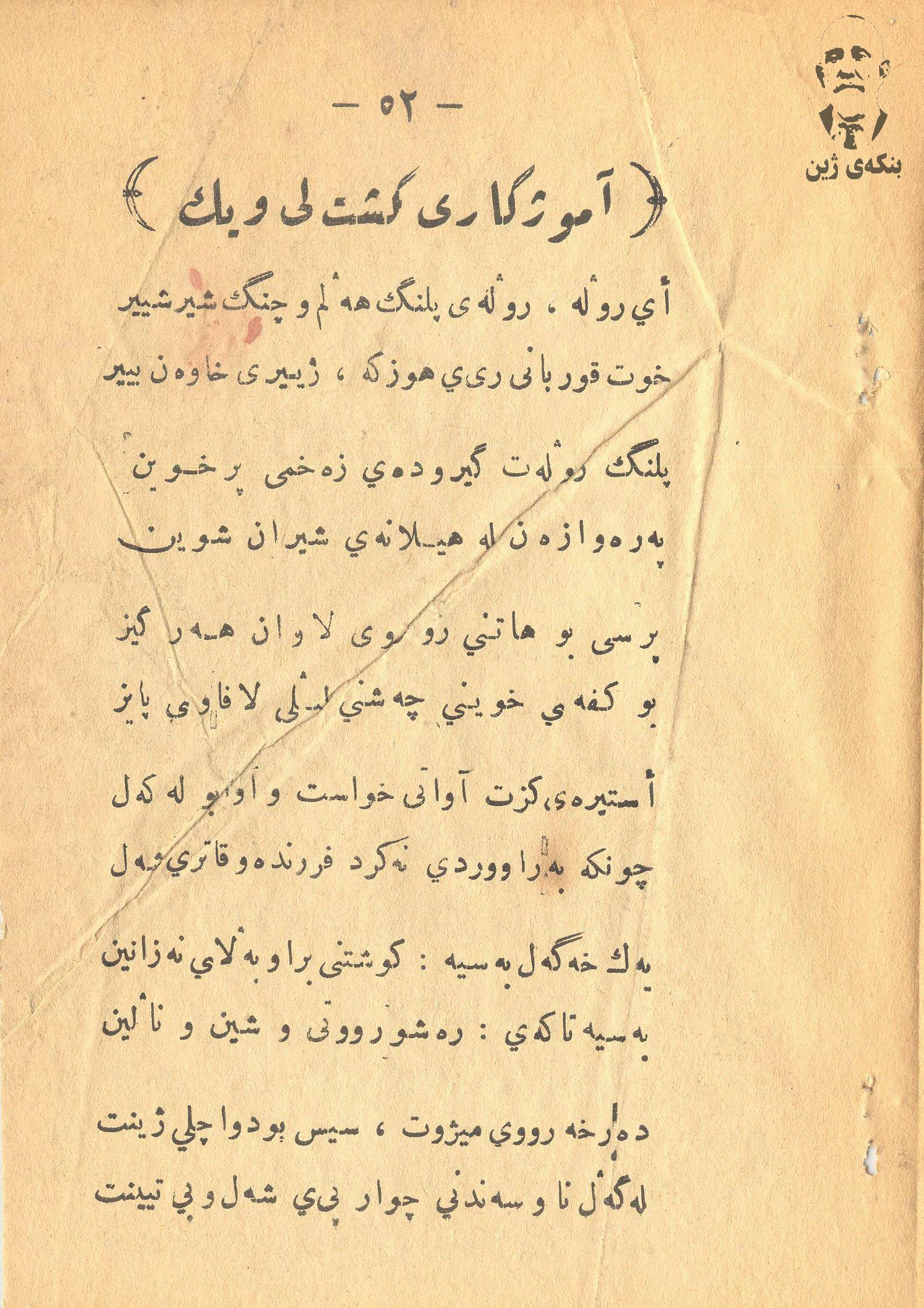}}
	\caption[A sample page from the book titled 'Deste Gullî Lawane' published in 1939.]{\label{fig:introduction_sample} A sample page from the book titled 'Deste Gullî Lawane' published in 1939 (Zheen Center for Documentation and Research).}
\end{figure}

Over the course of centuries, human experience has produced invaluable treasures in the form of historical documents. Due to the large amount of work required for manual annotation and transcription of historical documents, many archives of historical documents remain inaccessible \cite{ataer2007matching}. Through digitization, these documents can be understood and protected efficiently and effectively. In this process, actual documents are systematically converted into digital records based on the precise recognition of characters in the original document \cite{yang2018recognition}. Because of the demand for maintaining and making historical documents available for research without damaging physical copies, many languages and regions started practicing and studying digitization and preservation of the digital reproduction of historical documents \cite{nguyen2017attempts}. According to \newcite{poncelas2020tool}, building Optical Character Recognition (OCR) that recognizes and extracts text from historical documents is a challenging task, and some unique sets of issues can affect the result of the model. Typeface inconsistency and bad-quality images are some examples of the challenges. Figure ~\ref{fig:introduction_sample} is a sample page with these challenges. As a result of these issues, most of the advanced OCR systems produce errors which is why researchers continue their efforts to find new methods to enhance the OCR engines to generate better output.

Initially, historical documents were painstakingly created by hand, leading to their restricted availability and limited distribution. However, the introduction of the printing press by Johannes Gutenberg in 1436 in Germany marked a significant milestone. The printing press, a mechanical device designed for printing high-volume publications, revolutionized the production of historical documents. This apparatus applies pressure on an inked surface, as depicted in Figure ~\ref{fig:guttenbergs_printing_press}. The printing press is widely recognized as one of the most remarkable accomplishments in history, facilitating the widespread dissemination and preservation of knowledge \cite{qania2012}.

As for the Kurdish press history, it is about one century old, and the devices used for printing were hugely different from what we have today. The devices underwent many changes and improvements until we reached what we have today.

Publications printed with the printing press have various issues. One of them is the lack of standard font for writing, the use of many Arabic styles, and on top of them, all the books need to be in better shape as they are very fragile and damaged and there are many noticeable marks on them.

A few OCR systems currently support the Kurdish language, for example, the one by \newcite{idrees2021exploiting}. Still, they cannot recognize these old publications due to the abovementioned issues. As for the old publications, some works have been done for the other languages that we go through in the literature review chapter.

This study focuses on enhancing an existing OCR system for the Kurdish language so we can recognize and extract text from historical Kurdish documents, which makes the related documents ready for further processing.

\subsection{Printing Press in Iraq and Iraqi Kurdistan}

According to \newcite{hassanpour1992nationalism} printing presses first appeared in the Kurdish towns of the Ottoman Empire in the late 1860s. They were all established, owned, and operated by the government for printing in Turkish. A number of large cities with small Kurdish populations on the outskirts of Kurdistan had an earlier start - Mosul began with a press established by Dominican missionaries in 1856, followed by a government press in 1881, while Erzurum had a government press as early as 1865-66. The age of printing in Kurdish began in a different place than in Kurdistan. All Kurdish books and periodicals published in Kurdish during the Ottoman period were printed outside Kurdistan in Cairo, Istanbul, and Baghdad. This was because (a) except for the Dominican press in Mosul, all other presses were owned by a government not interested in Kurdish publications, (b) publishing was begun by Kurdish nationalists who were mostly in exile in Istanbul and other large cities, and (c) censorship was less effective in Cairo and Istanbul. The earliest record of a printing press owned by Kurds is "Matba'a Kurdistan `Ilmiya" (Scientific Kurdistan Press), established by Faraj-Allah Zaki al-Kurdi in Cairo, Egypt. They printed in Arabic, but there is no evidence of printing in Kurdish with this press.

\begin{figure}
	\centering
	\frame{\includegraphics[scale=0.5]{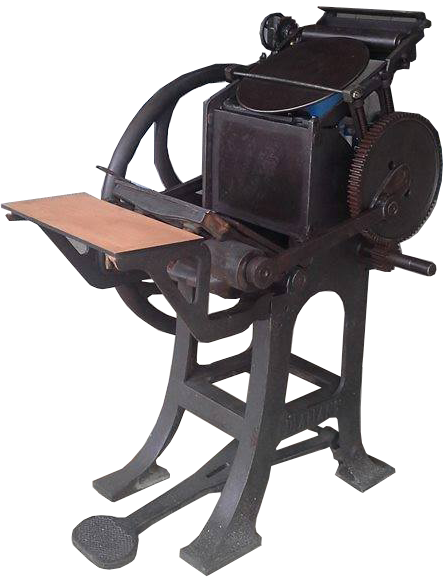}}
	\caption{\label{fig:first_erbil_press} Hussein Huzni's First Press (Dr. Kurdistan Mukryani's Archive)}
\end{figure}

As for Iraq, Giw Mukiryani, the nationalist language former journalist and printer, has claimed in several of his books that the first Kurdish press was founded in 1915 in Aleppo (now Syria) by his brother Husen Huzni Mukriyani, who had printed several books and journals before moving the press to Rawandiz in 1925, Figure ~\ref{fig:first_erbil_press} shows the printing press by Husen Huzni Mukriyani. Many Kurds have uncritically accepted this claim. There is strong evidence that this press did not exist. According to a notice in Diyari Kurdistan, "Huzni was in the process of founding" a printing press in Aleppo in 1925 and was asking the Kurds, through this magazine, to pre-purchase a book on Kurdish history and society that he was going to publish.  

The first press in Iraqi Kurdistan was set up in Sulaymania by the Mandate authorities in 1920. It was an old hand-operated letterpress called Chapkhanay or Matba'a Hukumat (Government Press). It printed six books, 118 issues of the weekly Peskawtin, 14 Issues of Bangi Kurdistan, and 16 issues of Roji Kurdistan between 1920 and 1923. The press and two schools held significant intellectual value for the autonomous government of Shaikh Mahmud, who frequently rebelled against Baghdad and proclaimed himself the King of Kurdistan. Like the newly formed Kurdish state, the press experienced a tumultuous existence. After Baghdad gained control over the region, the press came under the operation of the Municipality. It was subsequently renamed Matba'a Baladiya or Chapkhanay Sharawani (1925) and began printing a government-sponsored weekly publication called Jiyanewe, along with several books.

Huzni Mukriyani bought an outdated printing press in Syria and carried it by mule to Rawandiz in 1926, naming his enterprise Matba'a Zari Kirmanji (Kurdish Tongue Press). Huzni and his brother Giw used the ancient instrument to print twenty-three books (i.e., 24.2\% of 95 Kurdish books published in Iraq by 1938) and one magazine Zari Kirmanji between 1926 and 1930. The press moved to Arbil, where it began publishing the weekly Runaji in October 1935. It was renamed Chapkhanay Kurdistan (Kurdistan Press) after Huzni passed away in 1947 and began being owned and operated by Giw.

Due to the inability to utilize the Municipality Press, Piramerd purchased a larger, albeit used hand-press with worn-out letter types. This press, known as Zhiyan Press, commenced operations in September 1937. It focused on printing the weekly publication 'Jin'. Consequently, by 1937, three Kurdish presses were active: two in Sulaymaniah and one in Arbil. These presses played a role in publishing weekly materials for the Kurdish community.

Kurdish printing developed in Iraq at a time when the language was undergoing conscious and rapid codification, and it could play, as such, a very significant role in the culturalization process. One aspect of codification that depended on printing was orthographic reform.	The successful reform of the alphabet required the casting of new letters and the use of diacritical signs.

In Baghdad, where 58.6\% of all Kurdish books were published before 1958, the letters {\small{\RL{گ}}}, {\small{\RL{ژ}}}, {\small{\RL{چ}}}, and {\small{\RL{پ}}} representing the Kurdish phonemes \textit{p, {\c{c}}, j and g} (which do not exist in Arabic) were not found in most presses. Many books printed in the 1920s and 1930s lack these letters, which have been replaced partly or in whole by {\small{\RL{ک}}}, {\small{\RL{ز}}}, {\small{\RL{ج}}}, {\small{\RL{ب}}}, which stand for \textit{b,c,z,k}. Thus letter types had to be cast abroad, and printers were reluctant to invest in such an unprofitable venture. In Kurdistan, the press owners were financially unable to replace the worn-out letters, let alone cast new ones. Mukiryani recalls that when 
diacritics {\includegraphics[height=\fontcharht\font`\B]{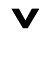}} and {\includegraphics[height=\fontcharht\font`\B]{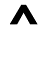}} were becoming fashionable in earlier decades, he and his brother could not use them in printing for lack of letter types. They used, instead, the numbers {\includegraphics[height=\fontcharht\font`\B]{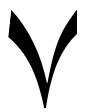}} 7 and {\includegraphics[height=\fontcharht\font`\B]{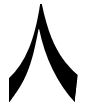}} 8, which made the text cumbersome, because they were full-size typefaces and could be placed only beside the letter they were to modify rather than over it. They did not have enough '7' and '8' types to set more than a few book pages. Huzni was forced to make woodcuts that were challenging to use. 

Unable to add diacritics, printers, and writers had to double letters, e.g., {\small{\RL{رر}}} for {\^r} and {\small{\RL{وو}}} for {\^u}. Letter-type problems were solved in the 1970s when typecasting became possible in Iraq, and several presses in Baghdad and other towns could provide the letters at a lower cost. Even more difficult than reforming the Arabic orthography was the adaptation of the Roman alphabet for Kurdish. No printing press in Baghdad had, or was willing to acquire, Roman letters with diacritics in 1957, when Jamal Nebez published his booklet, Nusini Kurdi be Latini' writing Kurdish in Latin [letters]'. When the best-equipped printer in the capital, the Ma'arif press, finally printed it, the writer had to manually add diacritical marks for seven letters ({\c{c}},{ {\"{h}}, {\l}, {r},{\c{s}} {\"u}, {\"{x}}) in all printed copies, each of which required hundreds of additions.

\subsection{Challenges in Historical Documents}

It is crucial to understand the defects and degradations in historical documents clearly. Figure ~\ref{fig:degraded_defects} illustrates these defects' commonly encountered degraded documents.

\begin{figure}[ht]
	\centering
	\frame{\includegraphics[scale=0.5]{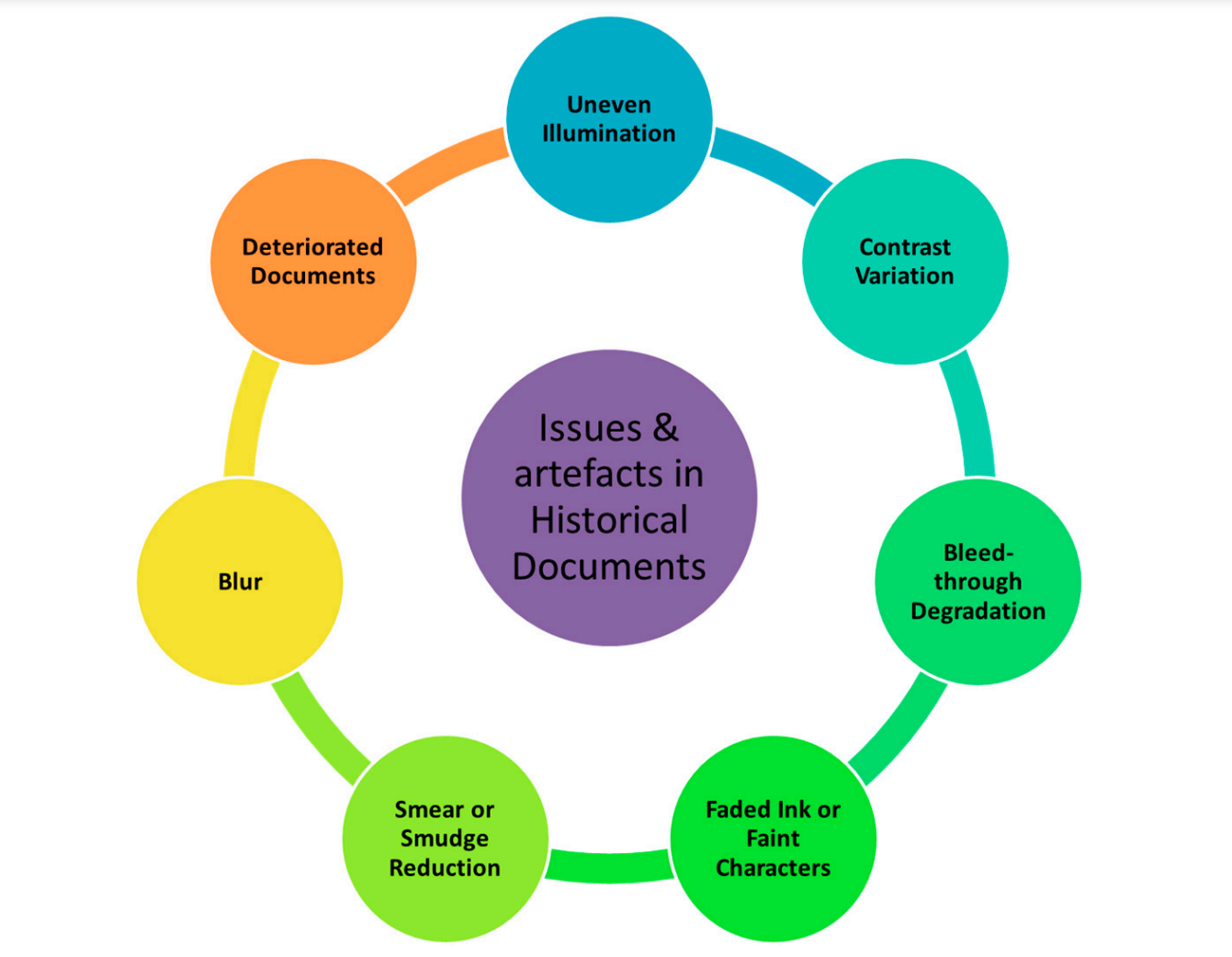}}
	\caption[Most frequently seen degraded defects in historical documents]{\label{fig:degraded_defects} Most frequently seen degraded defects in historical documents \cite{sulaiman2019degraded}} 
\end{figure}

\subsubsection{Uneven Illumination}
Uneven illumination in optical imaging degrades light microscopy images due to the diminishing of incident light along its path caused by particle spreading in the media. This leads to difficulties in document image analysis, especially in character recognition using OCR. Background objects, fluorescence overlays, and light scattering contribute to uneven illumination. This issue negatively affects efficient document recognition and can be seen in historical document examples. The typical OCR process of converting grayscale images to binary and extracting text is hindered by uneven illumination. Figure ~\ref{fig:uneven_illumination} shows an example of uneven illumination in historical documents.

\begin{figure}[ht]
	\centering
	\frame{\includegraphics[scale=0.5]{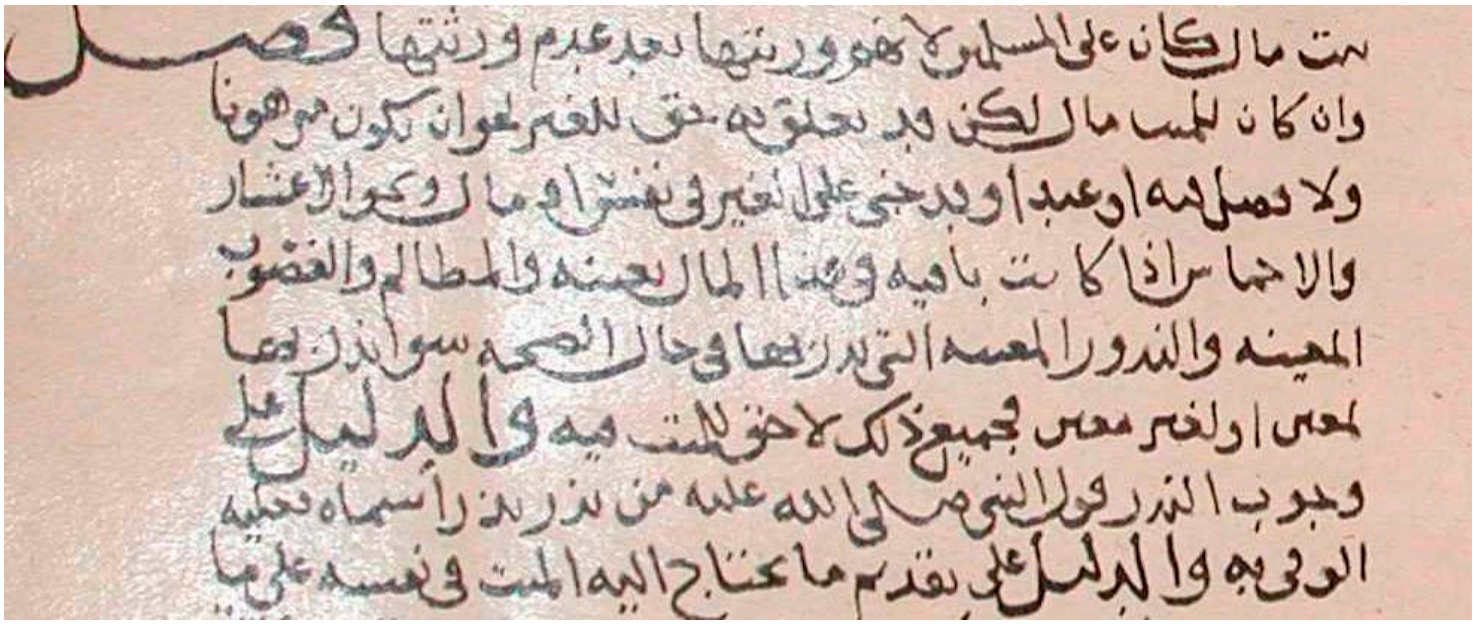}}
	\caption[Uneven illumination in handwritten historical document from Arabic databases]{\label{fig:uneven_illumination} Uneven illumination in handwritten historical document from Arabic databases \cite{sulaiman2019degraded}} 
\end{figure}

\subsubsection{Contrast Variation}
Contrast refers to the variation in brightness within an image. It primarily represents the differences between high-intensity and low-intensity pixels or the disparities between object pixels and background pixels. Factors like noise, sunlight, illumination, and occlusion can cause non-linear and expressive variations in contrast. These variations pose challenges for document image analysis algorithms, particularly in applying traditional threshold-based methods to distinguish foreground text from the background in historical and handwritten documents. To address this issue, image enhancement techniques can be employed prior to image binarization. Figure ~\ref{fig:contrast_variation} shows an example of contrast variation in handwritten historical documents.

\begin{figure}[ht]
	\centering
	\frame{\includegraphics[scale=0.5]{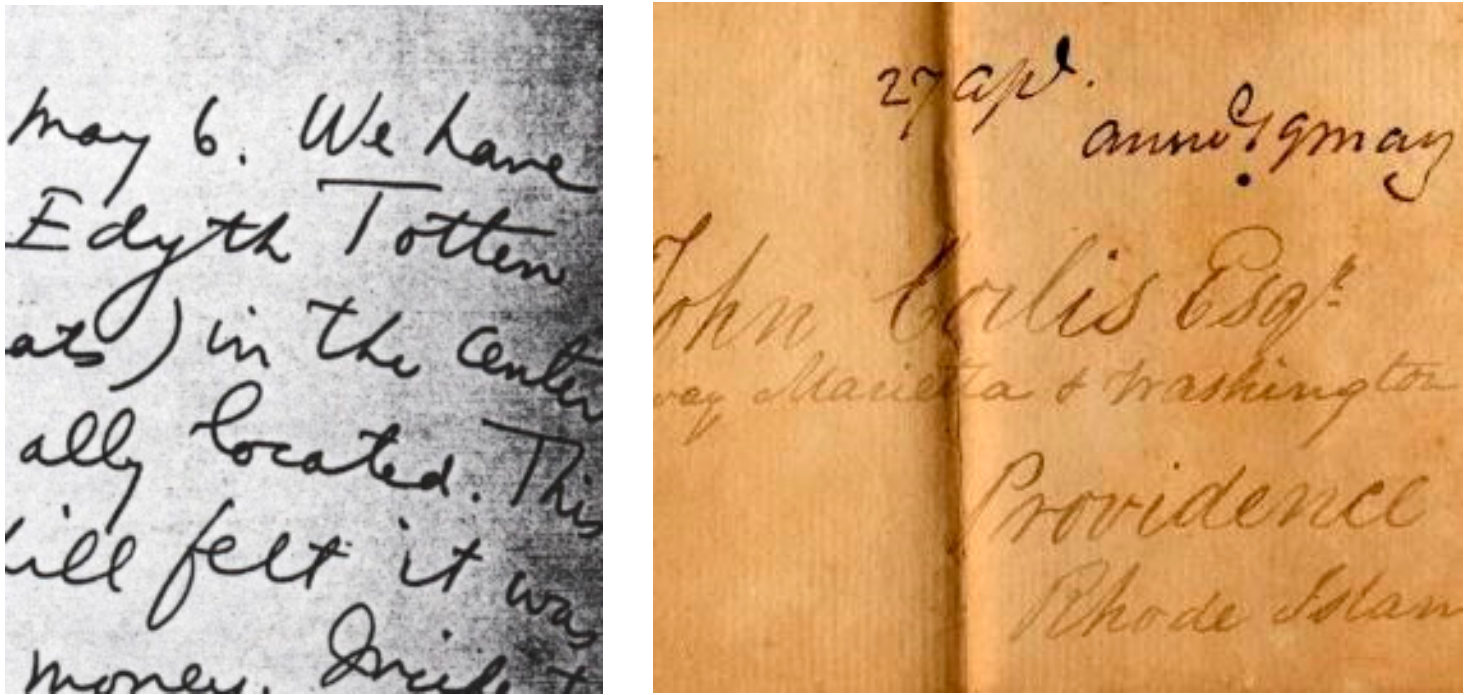}}
	\caption[Degraded document image showing variation of contrast]{\label{fig:contrast_variation} Degraded document image showing variation of contrast \cite{sulaiman2019degraded}} 
\end{figure}

\subsubsection{Bleed-Through Degradation}
Bleed-through, also known as ink bleeding, is a phenomenon where ink from one side of a paper document transfers to the other side, making the text illegible. This poses a significant challenge in document binarization, which aims to separate the foreground text from the background. Researchers addressing this issue have faced two major challenges: limited access to high-resolution degraded documents and the difficulty of quantitatively analyzing restoration outcomes without ground truth data. Solutions involve generating degraded images based on known ground truth or using known degraded images as references. Performance analysis can still be conducted by evaluating the impact of restoration on subsequent processes, such as OCR. Figure ~\ref{fig:bleed_through_degradation} shows an example of ink-bleed degradation.

\begin{figure}[ht]
	\centering
	\frame{\includegraphics[scale=0.5]{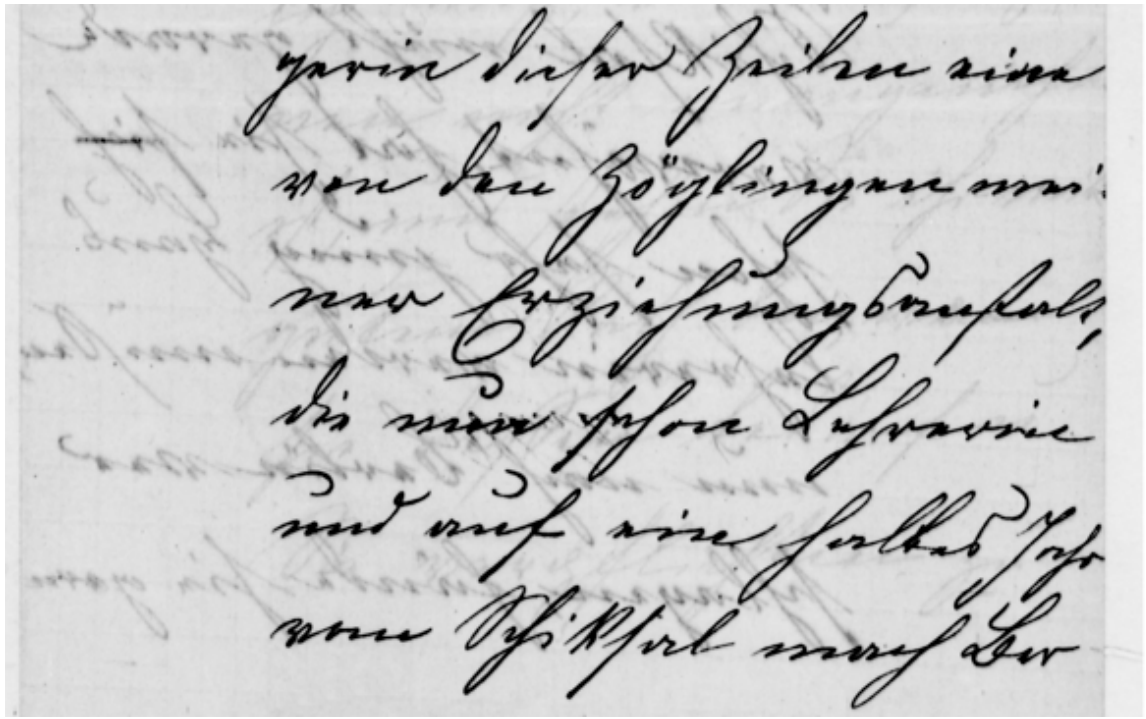}}
	\caption[Example of ink-bleed degradation in handwritten]{\label{fig:bleed_through_degradation} Example of ink-bleed degradation in handwritten documents \cite{sulaiman2019degraded}} 
\end{figure}

\subsubsection{Faded Ink or Faint Characters}
There is a strong interest in digitizing official organizational papers for historical, public, and political purposes. However, typewritten documents present challenges for recognition. The intensity of each character can vary compared to the surrounding glyphs due to factors like the typewriter key's striking head and the force applied during typing. Additionally, many typewritten documents exist only as carbon copies, resulting in blurry text due to the pressure required for clear imprints on both the original and carbon paper. Historical typewritten documents also face issues such as aging, tears, stains, rust, punch holes, disintegration, and discoloration. Figure ~\ref{fig:degraded_defects} shows instances of scanned historical documents with faded ink degradation.

\begin{figure}[ht]
	\centering
	\frame{\includegraphics[scale=0.5]{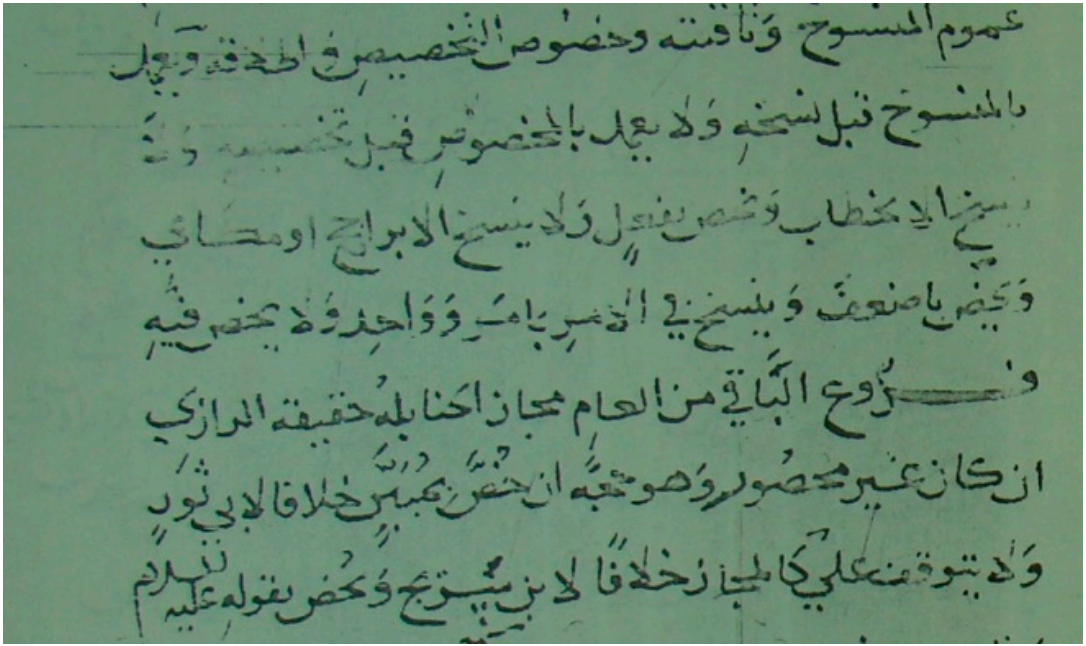}}
	\caption[Image showing faded degradation]{\label{fig:faded_ink_or_faint_characters} Image showing faded degradation \cite{sulaiman2019degraded}} 
\end{figure}

\subsubsection{Smear or Show Through}
After the digitization of documents, new challenges arise in the form of noise and low-resolution components that negatively affect the document's visual appearance. Historical documents can suffer from various types of degradation, introduced over time and with different characteristics. One prominent issue is show-through, where ink impressions from one side of the paper appear on the other side, making the document difficult to read. Restoration techniques are necessary to make these documents easily readable. Removing show-through improves readability and reduces image compression time, allowing for faster downloading over the internet. Figure ~\ref{fig:smear_or_show_through} Shows an example of such degradation.

\begin{figure}[ht]
	\centering
	\frame{\includegraphics[scale=0.5]{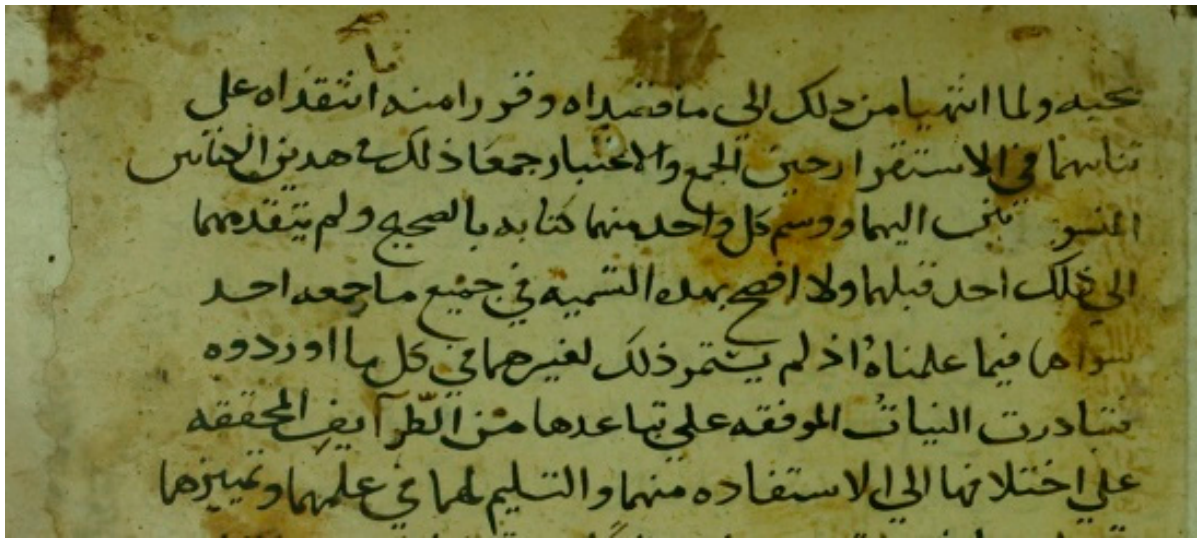}}
	\caption[Example showing degraded document with smear/show-through effects]{\label{fig:smear_or_show_through} Example showing degraded document with smear/show-through effects \cite{sulaiman2019degraded}} 
\end{figure}

\subsubsection{Blur}
Regarding document degradation, two types of blurring appear in documents: Motion blur and out-of-focus blur. In general, motion blur artifacts are caused by the relative speed between the camera and the object or a sudden rapid camera movement. In contrast, out-of-focus blur occurs when the light fails to converge in the image. In order to fix the blur issue, the research topics as of late have turned towards the tools for assessing blur in document images to figure out the accuracy of the OCR, hence providing the required response to the user to help them obtain new images in the hopes of getting better OCR outcomes. Some instances of blurring issues in degraded documents are displayed in Figure ~\ref{fig:blur}.

\begin{figure}[ht]
	\centering
	\frame{\includegraphics[scale=0.5]{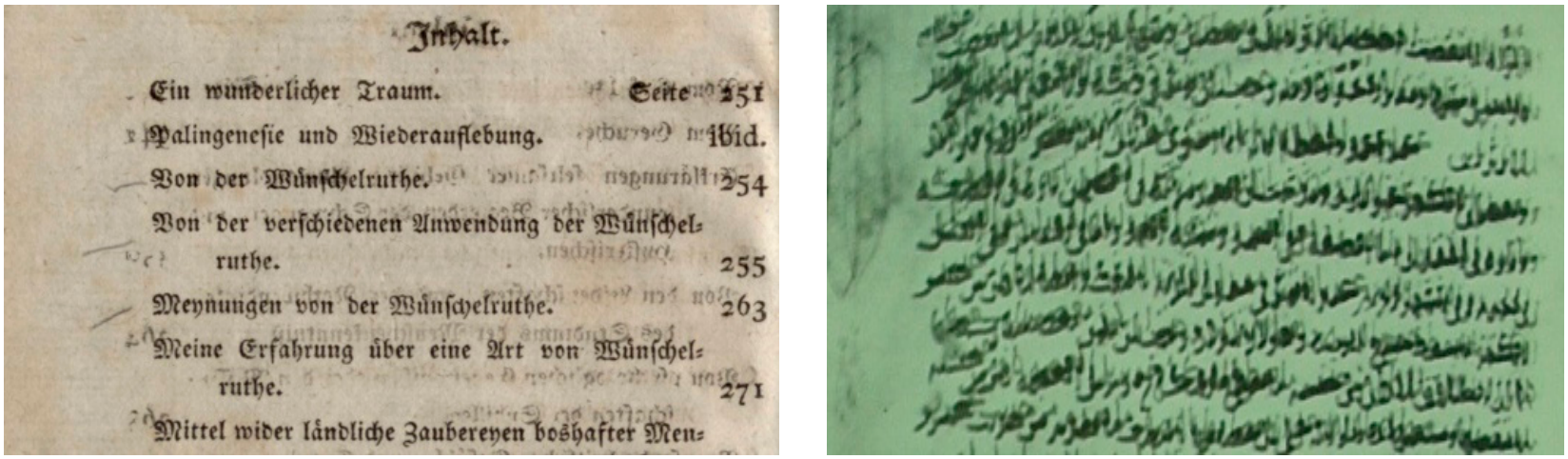}}
	\caption[Example showing degraded document with blurring effect]{\label{fig:blur} Example showing degraded document with blurring effect \cite{sulaiman2019degraded}} 
\end{figure}

\subsubsection{Thin or Weak Text}
Historical documents often contain thin or weak text, typically written with ink or paint. Over time, the ink used in these documents may shrink and degrade, making the text difficult to read. Additionally, using low-quality ink and paper can contribute to the appearance of thin or weak text, posing challenges for accurate text extraction through binarization methods. Recent research in prehistoric document image analysis has focused on addressing these challenges. Enhancement and binarization algorithms have been developed to improve the quality of thin or weak text in historical documents. Subsequent phases, such as skew detection, recognition, and page or line segmentation, have been created to process the binarized data. Figure ~\ref{fig:think_or_weak_text} shows an example of thin or weak text. 

\begin{figure}[ht]
	\centering
	\frame{\includegraphics[scale=0.5]{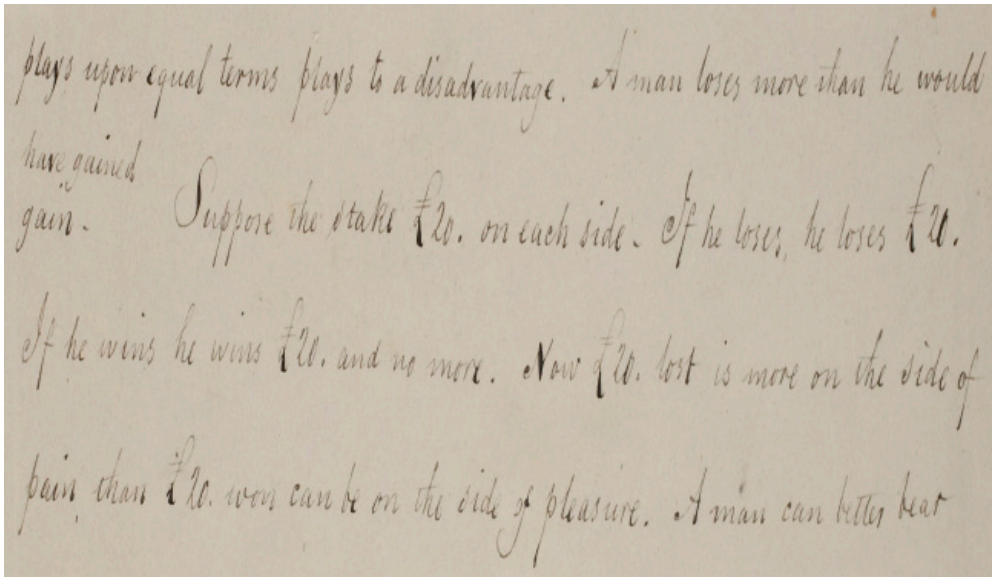}}
	\caption[Example showing thin or weak text from old document]{\label{fig:think_or_weak_text} Example showing thin or weak text from old document \cite{sulaiman2019degraded}} 
\end{figure}

\subsubsection{Deteriorated Documents}
Original paper-based documents can encompass various media types (such as ink, graphite, and watercolor) and formats (such as rolled maps, spreadsheets, and record books). These documents hold significant importance due to their informational, evidential, associational, and intrinsic values. The evidential value of a document, particularly in historical, legal, or scientific contexts, relies on preserving the original condition of the media, substrate, format, and images without significant alterations or deterioration. However, documents can face deterioration, loss, and damage not only through actual use but also due to factors like poor storage, handling, environmental conditions, and inherent instability. Environmental factors, especially for inherently unstable documents, can cause severe damage and deterioration. Figure ~\ref{fig:deteriorated_documents} shows an example of a deteriorated document.

\begin{figure}[ht]
	\centering
	\frame{\includegraphics[scale=0.5]{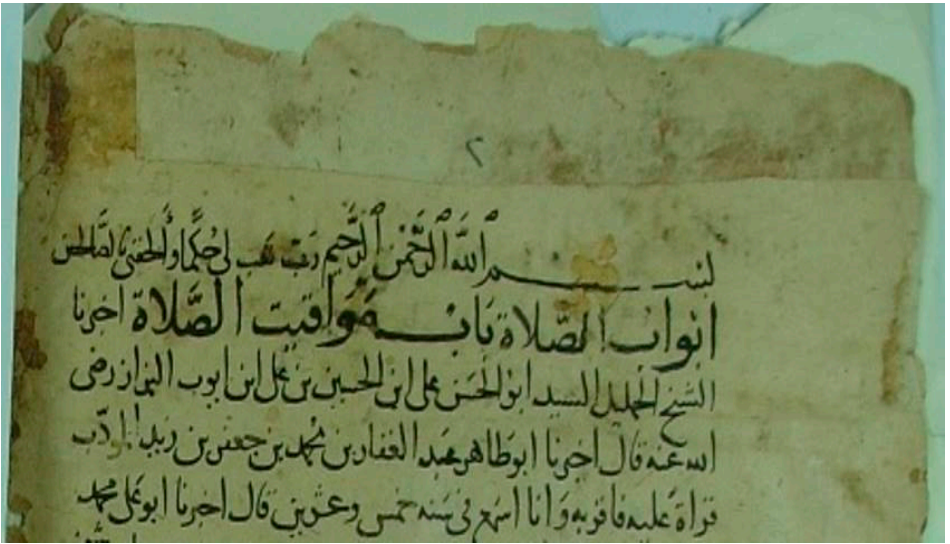}}
	\caption[An example of deteriorated document]{\label{fig:deteriorated_documents} An example of deteriorated document \cite{sulaiman2019degraded}} 
\end{figure}

\subsection{Kurdish Language}
Kurdish refers to various dialects in the region, encompassing Iran, Iraq, Turkey, and Syria. Nevertheless, Kurds have resided in additional countries, including Armenia, Lebanon, Egypt, and others, for several centuries. Additionally, they have substantial diaspora communities in various European countries and North America \cite{hassani2016automatic}. The precise number of speakers for this language remains uncertain, with varying reports suggesting a population between 19 million to 28 million \cite{hassani2016automatic}. Scholars often describe Kurdish as a dialect continuum, wherein language intelligibility can vary across different regions. Generally, Kurdish is recognized to encompass three primary dialects: Northern Kurdish (Kurmanji), Central Kurdish (Sorani), and Southern Kurdish \cite{ahmadi2022leveraging}. Kurdish utilizes four different scripts for writing, including modified Persian/Arabic, Latin, Yekgirtû (unified), and Cyrillic. The popularity and usage of these scripts vary depending on geographical and geopolitical factors \cite{hassani2016automatic}.

Sorani is commonly written using an adapted Persian/Arabic script with a cursive style, following a right-to-left (RTL) direction. See Figure ~\ref{fig:arabic_alphabet} for Arabic Alphabets, Figure ~\ref{fig:persian_alphabet} for Perian Alphabet and Figure ~\ref{fig:kurdish_alphabet} for Kurdish alphabet Persian/Arabic script. On the other hand, Kurmanji predominantly employs the Latin script for writing, except in the Kurdistan Region of Iraq and Kurdish areas of Syria, where the same script as Sorani is utilized \cite{idrees2021exploiting}.

Sorani primarily employs a modified Persian/Arabic script, while Zazaki mainly utilizes the Latin script. Gorani (Hawrami), on the other hand, is primarily written in a modified Persian/Arabic script. It is worth noting that the term "mainly" is used because there are significant exceptions in the usage of these scripts, particularly with regard to the Latin and modified Persian/Arabic scripts \cite{hassani2016automatic}.

\begin{figure}[ht]
	\centering
	\includegraphics[scale=0.60]{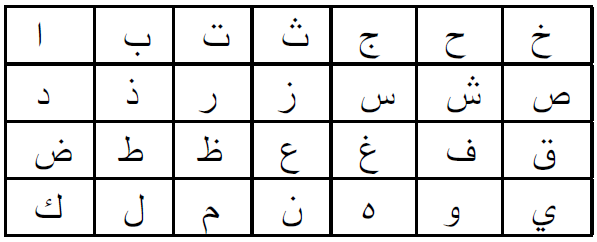}
	\caption{\label{fig:arabic_alphabet} Arabic alphabet}
\end{figure}

\begin{figure}[ht]
	\centering
	\includegraphics[scale=0.60]{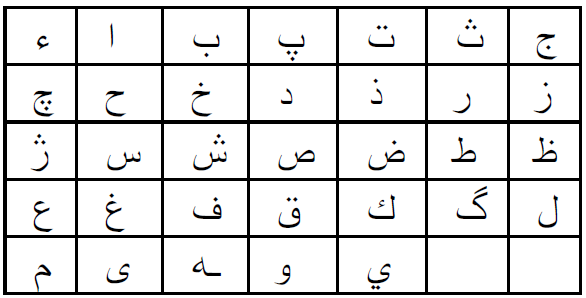}
	\caption{\label{fig:persian_alphabet} Persian alphabet}
\end{figure}

\begin{figure}[ht]
	\centering
	\includegraphics[scale=0.60]{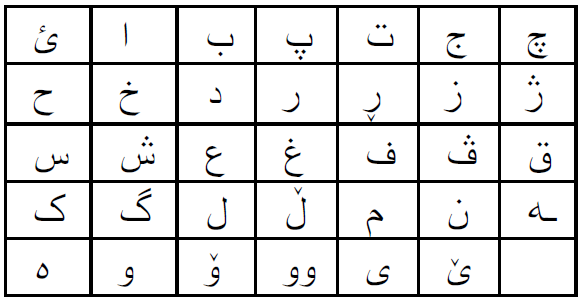}
	\caption{\label{fig:kurdish_alphabet} Kurdish alphabet}
\end{figure}
	
The rest of the paper is organized as follows. Section~\ref{Sec-Work} reviews the literature of OCR for historical documents for different languages. Section~\ref{Sec-Method} presents the method that the research follows. We provide the results and discuss the outcome in Section~\ref{sec-erd}. Finally, Section~\ref{Sec-Conc}  concludes the study, summarizes the findings, and introduces the opportunities for future work.

\section{Related work}
\label{Sec-Work}
This section reviews the literature by focusing on machine-typed historical documents. To the best of our knowledge, currently, there is no OCR system that can accurately extract text from old Kurdish publications written in Arabic-Persian script. Therefore, we concentrate on the related work for other languages.

\subsection{Arabic/Persian}

It is difficult to implement an Ottoman character recognition system according to \newcite{ozturk2000multifont}. There are insufficient studies in this field. Therefore, they developed a model using artificial neural networks using 28 different Ottoman machine-printed documents in order to develop an OCR that will recognize different fonts. Three Ottoman newspapers were used to prepare their data. For documents with a trained font, the accuracy was 95\%, while for documents with an unknown font, it was 70\%.

According to \newcite{ataer2007matching}, it may not be possible to obtain satisfactory results using character recognition-based systems due to the characteristics of Ottoman documents. Moreover, it is desirable to store the documents as images, since the documents may contain important drawings, especially signatures. The author viewed Ottoman words as images and proposed a matching technique to solve the problem because of these reasons. According to the author, the bag-of-visterm approach was shown to be successful in classifying objects and scenes, which is why he adopted the same approach for matching word images. Using vector quantization of Scale-Invariant Feature Transform (SIFT) descriptors, word images were represented by sets of visual terms. By comparing the visual terms' distributions, similar words are then matched. Over 10,000 words were included in the printed and handwritten documents used in the experiments. In the experiment, the highest accuracy was 91\% and the lowest accuracy was 30\%.

\newcite{kilic2008multifont} developed an OCR system specifically designed for Ottoman script segmentation, normalization, edge detection, and recognition. The Ottoman characters were categorized into four distinct forms based on their position within a word: beginning, middle, end, and isolated form. Images of printed papers containing Ottoman script were used for data acquisition. The process involved segmentation and normalization of the images, followed by edge detection using Cellular Neural Networks for feature extraction. Subsequently, a Support Vector Machine (SVM) was employed to accurately identify these multi-font Ottoman characters. The SVM training involved the utilization of Polynomial (linear and quadratic) and Gaussian Radial Basis Function kernels. The proposed recognition system achieved an impressive accuracy rate of 87.32 percent for character classification.

\newcite{shafii2014optical} proposed a new technique in two important preprocessing steps, skew detection and page segmentation, after reviewing the existing technology. Instead of utilizing the usual practice of segmenting characters, they suggested segmenting subwords to avoid challenges with segmentation due to Persian script's highly cursive nature. Feature extraction was implemented using a hybrid scheme that combines three commonly used methods before being classified using a nonparametric method. Based on their experimental tests on a library of 500 words, they were able to recognize 97\% of the words.

Due to the challenges of the Arabic heritage collection, which consists of early prints and manuscripts, it is difficult to extract text from its documents. To address these problems, \newcite{stahlberg2016qatip} developed a system called QATIP (QCRI Qatar Computing Research Institute Arabic Text Image Processing) to OCR these kind of documents. A sophisticated text-to-image binarization technique was used in conjunction with Kaldi, which was originally designed for speech recognition. This paper contributed two major areas, one involving the creation of both a graphical user interface for users as well as API endpoints for integration and the other new approaches to model language and ligatures. After testing the system, they found out that the newly proposed technique for language modeling and ligature modeling was highly successful. The accuracy of the system was 37.5\% WER 12.6\% CER for early books.

In order to recognize Ottoman-Turkish characters, \newcite{dougru2016ottoman} used Tesseract optical character recognition system. In addition, various transcription methods have been developed from Ottoman Turkish to Latin. Optical character recognition could not recognize certain Ottoman-Turkish characters. As a result, Ottoman-Turkish keyboards were developed to facilitate the writing of unrecognized characters using Ottoman-Turkish alphabets. For the transcription process, dictionary tables were used. This resulted in an increase in the success rate of transcription when enrichment data was included in the dictionary tables. Therefore, an application was developed to enrich dictionary tables with data.  The recognition rates for the first two pages of an Ottoman book was between 75.88\% - 77.38\%. Based on the results of the author's experiments, he concludes that recognition rates could vary based on quality, style, and printed or handwritten documents or images. High quality and printed images can be recognized with a 100\% accuracy rate, while handwritten and low-quality documents or images cannot be recognized by optical character recognition. It is therefore necessary to write these kinds of documents or images again in Ottoman-Turkish.

Analytical based approach for cursive scripts such as Arabic can be very challenging, especially for segmentation, because of the frequent overlapping between the characters. Because of that \newcite{nashwan2017holistic} proposed a segmentation-based holistic approach to solve this issue. Since we deal with the entire word as a single unit in the holistic approach, this will improve the error rate for cursive scripts. But on the other hand, it will require computation complexity especially if the application has a huge vocabulary. In their view, their holistic approach, based on Discrete Cosine Transforms (DCTs) and local block features, will be computationally efficient. In addition, they developed a method for reducing the length of the lexicon by clustering words that have similar shapes. The proposed system was tested on a wide range of datasets, and it was found to have a 47.8\% WRR accuracy, and it increased to 65.7\% WRR when considering the top-10 hypotheses.

By employing deep convolutional neural networks, \newcite{kuccukcsahin2019design} devised an offline OCR system that demonstrates the ability to recognize Ottoman characters. The proposed methodology encompasses multiple stages, including image processing, image digitization, character segmentation, adaptation of inputs for the network, network training, recognition, and evaluation of outcomes. To create a character dataset, text images of varying lengths were segmented from diverse samples of Ottoman literature obtained from the Turkish National Library's digital repository. Two convolutional neural networks of differing complexity were trained using the generated character dataset, and the correlation between recognition rates and network complexity was examined. The dataset's features were extracted through the Histogram of Oriented Gradients and Principal Component Analysis techniques, while classification of Ottoman characters was achieved using the widely employed k-Nearest Neighbor Algorithm and Support Vector Machines. Results from the conducted analyses revealed that both networks exhibit recognition rates comparable to traditional classifiers; however, the more intricate deep neural network outperformed others in terms of accuracy and loss. After 100 epochs, the most accurate model achieved an impressive accuracy of 97.58 percent.

\newcite{dolek2021ottoman} presented an OCR tool developed for printed Ottoman documents in Naskh font. The tool was developed using a deep learning model trained with data sets containing both original and synthetic documents. The model was compared with free and open-source OCR engines using a test dataset comprising 21 pages of original documents. In terms of accuracy rates, their model outperformed the other tools with 88.64\% raw, 95.92\% normalized, and 97.18\% joined. Additionally, their model achieved an accuracy rate of 58 percent for word recognition, which is the only rate above 50 percent among the OCR tools that were compared.

\subsection{Chinese/Japanese}

Historical Chinese characters have posed one of the greatest challenges in pattern recognition in the past. This is due to their large character set and various writing styles. To address this issue \newcite{li2014historical} proposed a method of recognizing historical Chinese characters by incorporating STM into an MQDF classifier. The experiment was conducted on historical documents from Dunhuang and traditional Chinese fonts. The optimal selection of parameters was selected after testing many different parameters. They conducted two separate sets of experiments. An experiment using printed traditional Chinese characters was conducted as part of the first set of experiments. For the second experiment, samples taken from historical Chinese documents were used to perform the experiments. In addition, the method may be improved by introducing non-linear transfer and integrating it with other approaches. Furthermore, the system was tested with a variety of features and classifiers. The results of experiments suggest that supervised STMs may improve the generalization of classifiers. As a result of the results, the error rate was reduced by a considerable amount and the method showed significant potential. For example, it is possible to reduce the error rate of one of the tested documents by 60\% by tagging only 10\% of the samples with labels.

The lack of labeled training samples makes recognition of historical Chinese characters very challenging. Therefore, \newcite{feng2015gaussian} proposed a non-linear Style Transfer Mapping (STM) model based on Gaussian Process (GP-STM), which extends the traditional linear STM model. By using GP-STM, existing printed samples of Chinese characters were used to recognize historical Chinese characters. To prepare the GP-STM framework, the researchers compared a number of methods for extracting features, trained a Modified Quadratic Discriminant Function (MQDF) classifier on examples of Chinese characters printed on paper, and then applied the model to historical documents from Dunhuang. The impact of different kernels and parameters was measured, in addition to the impact of the number of training samples. In the experiments, the results indicate that the GP-STM is capable of achieving an accuracy of 57.5\%, an improvement of over 8\% over the STM.

It is difficult to recognize Chinese characters directly by using classical methods when they appear in historical documents since they can be categorized into more than 8000 different categories. Due to the lack of well-labeled data, deep learning based methods are unable to recognize them. The authors of \newcite{yang2018recognition} presented a historical Chinese text recognition algorithm based on data that was labeled at the page level without aligning each line of text. In order to reduce the influence of misalignment between text line images and labels, they proposed Adaptive Gradient Gate (AGG). The proposed text recognizer can reduce its error rate by over 35 percent with the help of AGG. Furthermore, they found that establishing an implicit language model using Convolutional Neural Networks (CNNs) and Connectionist Temporal Classification (CTC) is one of the key factors in achieving high recognition performance. With an accuracy rate of 94.64\%, the proposed system outperformed other optical character recognition systems.

Deep reinforcement learning has found successful applications across various fields. \newcite{sihang2020precise} presented an innovative approach, based on deep reinforcement learning, to enhance the F-measure score for Chinese character detection in historical documents. Their method introduced a novel fully convolutional network called fully convolutional network with position-sensitive Region-of-Interest (RoI) pooling FCPN. Unlike fixed-size character patches, this network could accommodate patches of varying sizes and incorporate positional information into action features. Additionally, they proposed a Dense Reward Function (DRF) that effectively rewarded different actions based on environmental conditions, thereby enhancing the decision-making capability of the agent. The method was designed to be applicable to the output of character-level or word-level text detectors, resulting in more precise outcomes. The effectiveness of their approach was demonstrated through its application to the Tripitaka Koreana in Han (TKH) and Multiple Tripitaka in Han (MTH) datasets, where a notable improvement was observed, achieving an Intersection over Union (IoU) of 0.8.

The introduction of ARCED by \newcite{ly2020attention} presents a novel attention-based row-column encode-decoder model for recognizing multiple text lines in images without requiring explicit line segmentation. The recognition system comprises three key components: a feature extractor, a row-column encoder, and a decoder. By adopting an attention-based seq2seq approach, the proposed model achieves significantly lower error rates compared to previous state-of-the-art methods for both single and multiple text line recognition. The encoder component leverages a row-column Bidirectional Long Short-Term Memory (BLSTM) network, enabling the capture of sequential order information in both horizontal and vertical dimensions. This contributes to further reducing error rates within the attention-based model. Additionally, a residual LSTM network utilizes all prior attention vectors to generate predictive distributions in the decoder, leading to improved accuracy. Training of the entire system is conducted using a cross-entropy loss function, utilizing only document images and ground-truth text. To evaluate the performance of ARCED, the Kana-PRMU dataset, comprising Japanese historical documents, is employed. Experimental results demonstrate that ARCED outperforms existing recognition methods. Specifically, when evaluated on the level 2 and level 3 subsets of the Kana-PRMU dataset, the proposed ARCED model achieves character error rates of 4.15\% and 12.69\% respectively. Future work aims to enhance ARCED's capabilities for recognizing entire Japanese document pages. Furthermore, incorporating a language model into ARCED is anticipated to further enhance its performance.

\subsection{Coptic}
According to \newcite{bulert2017optical} due to non-standard fonts and varying paper and font quality, OCR results may not be satisfactory when applied to historical texts. Further, historical texts are not transmitted in their entirety over time, but rather include gaps and fragments. As a result, automatic post-correction is more difficult when it comes to historical texts than when it comes to modern texts. Two tools were used to create recognition patterns (or models) specific to different languages and documents to recognize printed Coptic texts. Historically, Coptic was the last stage in the development of the pre-Arabic language that was indigenous to Egypt. In addition, it led to the creation of a rich and unique body of literature, including monastic texts, Gnostic texts, Manichaean texts, magical texts, and translations of the Bible and patristic texts. According to the researchers, Coptic texts possess properties that make them excellent candidates for computer-based reading. As a result of their limited number and the fact that most handwritten texts exhibit highly consistent forms, the characters can easily be identified.

\subsection{Greek}
A study published by \newcite{simistira2015recognition} investigated the performance of LSTM for inputting Greek polytonic script in OCR. Even though there are many Greek polytonic manuscripts, digitization of such documents has not been widely applied, and very little work has been done on the recognition of these scripts. In this study, they collected many diverse Greek polytonic script pages into a novel database, called Polyton-DB, containing 15,689 text lines of synthetic and authentic printed scripts, and conducted baseline experiments with LSTM networks. LSTM is shown to have an error rate between 5.51 and 14.68 percent (depending on the document) and is better than Tesseract and ABBYY FineReader, two well-known OCR engines.

It is not possible to recognize Greek characters in early printed Greek books using traditional character recognition techniques. Because the writing of the same or consecutive words does not permit character or word segmentation, the character or word cannot be segmented. To address this issue, \newcite{poulos2010ocr} has developed a novel OCR system combining image preprocessing with computational geometry. Their objective was to perform OCR digitization of a large collection of digitized Greek early printed books dating from the late 15th century to the mid-18th century. In this method, image processing is performed through the use of image binarization and enhancement, the creation of a convex polygon that represents the feature extraction of each font, and the development of training and identification procedures based on algorithms for intersecting convex polygons. Among the major advantages of this method was the ability to control the authentication of an image of a published document or a partial modification of it to a high degree of reliability. Therefore, the proposed system uses smart geometric practices to determine the classification of a candidate letter. According to experimental results, the proposed method yields positive and negative verification scores that are greater than 92\% correct.

\subsection{Latin}
\newcite{vamvakas2008complete} presented a complete OCR methodology for recognizing historical documents. It is possible to apply this methodology to both machine-printed and handwritten documents. Due to its ability to adjust depending on the type of documents that we wish to process, it does not require any knowledge of fonts or databases. Three steps were involved in the methodology: The first two involved creating a database for training based on a set of documents, and the third involved recognizing new documents. First, a pre-processing step that includes image binarization and enhancement takes place. In the second step, a top-down segmentation approach is used to detect text lines, words, and characters. A clustering scheme is then adopted to group characters of similar shapes. In this process, the user is able to interact at any time in order to correct errors in clustering and assign an ASCII label. Upon completion of this step, a database is created for the purpose of recognition. Lastly, in the third step, the above segmentation approach is applied to every new document image, while the recognition is performed using the character database that was created in the previous step.  Based on the results of the experiments, the model was found to be 83.66\% accurate. Efforts will be taken in the future to optimize the current recognition results by exploiting new approaches for segmentation and new types of features to optimize the current recognition results.

In typical OCR systems, binarization is a crucial preprocessing stage where the input image is transformed into a binary form by removing unwanted elements, resulting in a clean and binarized version for further processing. However, binarization is not always perfect, and artifacts introduced during this process can lead to the loss of important details, such as distorted or fragmented character shapes. Particularly in historical documents, which are prone to higher levels of noise and degradation, binarization methods tend to perform poorly, impeding the effectiveness of the overall recognition pipeline. To address this issue, \newcite{yousefi2015binarization} proposes an alternative approach that bypasses the traditional binarization step. They propose training a 1D LSTM network directly on gray-level text data. For their experiments, they curated a large dataset of historical Fraktur documents from publicly accessible online sources, which served as training and test sets for both grayscale and binary text lines. Additionally, to investigate the impact of resolution, they utilized sets of both low and high resolutions in their experiments. The results demonstrated the effectiveness of the 1D LSTM network compared to binarization. The network achieved significantly lower error rates, outperforming binarization by 24\% on the low-resolution set and 19\% on the high-resolution set. This approach offers a promising alternative by leveraging LSTM networks to directly process gray-level text data, bypassing the limitations and artifacts associated with traditional binarization methods. It proves particularly beneficial for historical documents and provides improved accuracy in OCR tasks.

According to \newcite{springmann2016automatic}, achieving accurate OCR results for historical printings requires training recognition models using diplomatic transcriptions, which are scarce and time-consuming resources. To overcome this challenge, the authors propose a novel method that avoids training separate models for each historical typeface. Instead, they employ mixed models initially trained on transcriptions from six printings spanning the years 1471 to 1686, encompassing various fonts. The results demonstrate that using mixed models yields character accuracy rates exceeding 90\% when evaluated on a separate test set comprising six additional printings from the same historical period. This finding suggests that the typography barrier can be overcome by expanding the training beyond a limited number of fonts to encompass a broader range of (similar) fonts used over time. The outputs of the mixed models serve as a starting point for further development using both fully automated methods, which employ the OCR results of mixed models as pseudo ground truth for training subsequent models, and semi-automated methods that require minimal manual transcriptions. In the absence of actual ground truth, the authors introduce two easily observable quantities that exhibit a strong correlation with the actual accuracy of each generated model during the training process. These quantities are the mean character confidence (C), determined by the OCR engine OCRopus, and the mean token lexicality (L), which measures the distance between OCR tokens and modern wordforms while accounting for historical spelling patterns. Through an ordinal classification scale, the authors determine the most effective model in recognition, taking into account the calculated C and L values. The results reveal that a wholly automatic method only marginally improves OCR results compared to the mixed model, whereas hand-corrected lines significantly enhance OCR accuracy, resulting in considerably lower character error rates. The objective of this approach is to minimize the need for extensive ground truth generation and to avoid relying solely on a pre-existing typographical model. By leveraging mixed models and incorporating manual corrections, the proposed method demonstrates advancements in OCR results for historical printings, offering a more efficient and effective approach to training recognition models.

\newcite{bukhari2017anyocr} introduced the "anyOCR" system, which focuses on the accurate digitization of historical archives. This system, being open-source, allows the research community to easily employ anyOCR for digitizing historical archives. It encompasses a comprehensive document processing pipeline that supports various stages, including layout analysis, OCR model training, text line prediction, and web applications for layout analysis and OCR error correction. One notable feature of anyOCR is its capability to handle contemporary images of documents with diverse layouts, ranging from simple to complex. Leveraging the power of LSTM networks, modern OCR systems enable text recognition. Moreover, anyOCR incorporates an unsupervised OCR training framework called anyOCRModel, which can be readily trained for any script and language. To address layout and OCR errors, anyOCR offers web applications with interactive tools. The anyLayoutEdit component enables users to rectify layout issues, while the anyOCREdit component allows for the correction of OCR errors. Additionally, the research community can access a Dockerized Virtual Machine (VM) that comes pre-installed with most of the essential components, facilitating easy setup and deployment. By providing these components and tools, anyOCR empowers the research community to utilize and enhance them according to their specific requirements. This collaborative approach encourages further refinement and advancements in the field of historical archive digitization.

\newcite{springmann2018ground} provided resources for historical OCR called the GT4HistOCR dataset, which consists of printed text line images accompanied by corresponding transcriptions. This dataset encompasses a total of 313,173 line pairs derived from incunabula spanning the 15th to the 19th centuries. It is made publicly available under the CC-BY 4.0 license, ensuring accessibility and usability. The GT4HistOCR dataset is particularly well-suited for training advanced recognition models in OCR software that utilize recurrent neural networks, specifically the LSTM architecture, such as Tesseract 4 or OCRopus. To assist researchers, the authors have also provided pretrained OCRopus models specifically tailored to subsets of the dataset. These pretrained models demonstrate impressive character accuracy rates of 95 percent for early printings and 98.5 percent for 19th-century Fraktur printings, showcasing their effectiveness even on unseen test cases.

According to \newcite{nunamaker2016tesseract}, historical document images must be accompanied by ground truth text for training an OCR system. However, this process typically requires linguistic experts to manually collect the ground truths, which can be time-consuming and labor-intensive. To address this challenge, the authors propose a framework that enables the autonomous generation of training data using labelled character images and a digital font, eliminating the need for manual data generation. In their approach, instead of using actual text from sample images as ground truth, the authors generate arbitrary and rule-based "meaningless" text for both the image and the corresponding ground truth text file. They also establish a correlation between the similarity of character samples in a subset and the performance of classification. This allows them to create upper- and lower-bound performance subsets for model generation using only the provided sample images. Surprisingly, their findings indicate that using more training samples does not necessarily improve model performance. Consequently, they focus on the case of using just one training sample per character. By training a Tesseract model with samples that maximize a dissimilarity metric for each character, the authors achieve a character recognition error rate of 15\% on a custom benchmark of 15th-century Latin documents. In contrast, when a traditional Tesseract-style model is trained using synthetically generated training images derived from real text, the character recognition error rate increases to 27\%. These results demonstrate the effectiveness of their approach in generating training data autonomously and improving the OCR performance for historical documents.

\newcite{koistinen2017improve} documented the efforts undertaken by the National Library of Finland (NLF) to enhance the quality of OCR for their historical Finnish newspaper collection spanning the years 1771 to 1910. To analyze this collection, a sample of 500,000 words from the Finnish language section was selected. The sample consisted of three parallel sections: a manually corrected ground truth version, an OCR version corrected using ABBYY FineReader version 7 or 8, and an ABBYY FineReader version 11-reOCR version. Utilizing the original page images and this sample, the researchers devised a re-OCR procedure using the open-source software Tesseract version 3.04.01. The findings reveal that their method surpassed the performance of ABBYY FineReader 7 or 8 by 27.48 percentage points and ABBYY FineReader 11 by 9.16 percentage points. At the word level, their method outperformed ABBYY FineReader 7 or 8 by 36.25 percent and ABBYY FineReader 11 by 20.14 percent. The recall and precision results for the re-OCRing process, measured at the word level, ranged between 0.69 and 0.71, surpassing the previous OCR process. Additionally, other metrics such as the ability of the morphological analyzer to recognize words and the rate of character accuracy demonstrated a significant improvement following the re-OCRing process.

\newcite{reul2018state} examined the performance of OCR on 19th-century Fraktur scripts using mixed models. These models are trained to recognize various fonts and typesets from previously unseen sources. The study outlines the training process employed to develop robust mixed OCR models and compares their performance to freely available models from popular open-source engines such as OCRopus and Tesseract, as well as to the most advanced commercial system, ABBYY. To evaluate a substantial amount of unknown information, the researchers utilized 19th-century data extracted from books, journals, and a dictionary. Through the experiment, they found that combining models with real data yielded better results compared to combining models with synthetic data. Notably, the OCR engine Calamari demonstrated superior performance compared to the other engines assessed. It achieved an average CER of less than 1 percent, a significant improvement over the CER exhibited by ABBYY.

According to \newcite{romanello2021optical}, commentaries have been a vital publication format in literary and textual studies for over a century, alongside critical editions and translations. However, the utilization of thousands of digitized historical commentaries, particularly those containing Greek text, has been challenging due to the limitations of OCR systems in terms of poor-quality results. In response to this, the researchers aimed to evaluate the performance of two OCR algorithms specifically designed for historical classical commentaries. The findings of their study revealed that the combination of Kraken and Ciaconna algorithms achieved a significantly lower CER compared to Tesseract/OCR-D (average CER of 7\% versus 13\% for Tesseract/OCR-D) in sections of commentaries containing high levels of polytonic Greek text. On the other hand, in sections predominantly composed of Latin script, Tesseract/OCR-D exhibited slightly higher accuracy than Kraken + Ciaconna (average CER of 8.2\% versus 8.2\%). Additionally, the study highlighted the availability of two resources. Pogretra is a substantial collection of training data and pre-trained models specifically designed for ancient Greek typefaces. On the other hand, GT4HistComment is a limited dataset that provides OCR ground truth specifically for 19th-century classical commentaries.

According to \newcite{skelbye2021ocr}, the use of deep CNN-LSTM hybrid neural net- works has proven to be effective in improving the accuracy of OCR models for various languages. In their study, the authors specifically examined the impact of these networks on OCR accuracy for Swedish historical newspapers. By employing the open source OCR engine Calamari, they developed a mixed deep CNN-LSTM hybrid model that surpassed previous models when applied to Swedish historical newspapers from the period between 1818 and 1848. Through their experiments using nineteenth-century Swedish newspaper text, they achieved a remarkable average Character Accuracy Rate (CAR) of 97.43 percent, establishing a new state-of-the-art benchmark in OCR performance.

Based on \newcite{aula2021improvement}, scanned documents can contain deteriorations acquired over time or as a result of outdated printing methods.  There are a variety of visual attributes that can be observed on these documents, such as variations in style and font, broken characters, varying levels of ink intensity, noise levels and damage caused by folding or ripping, among others. Modern OCR tools are unfavorable to many of these attributes, leading to failures in the recognition of characters. To improve the result of character recognition, they used image processing methods. Furthermore, common image quality characteristics of scanned historical documents with unidentifiable text were analyzed. For the purposes of this study, the open-source Tesseract software was used for optical character recognition. To prepare the historical documents for Tesseract, Gaussian lowpass filtering, Otsu's optimum thresholding method, and morphological operations were employed. The OCR output was evaluated based on the Precision and Recall classification method. It was found that the recall had improved by 63 percentage points and the precision by 18 percentage points. This study demonstrated that using image pre-processing methods to improve the readability of historical documents for the use of OCR tools has been effective.

According to \newcite{gilbey2021end}, it is noted that historical and contemporary printed documents often have extremely low resolutions, such as 60 dots per inch (dpi). While humans can still read these scans fairly easily, OCR systems encounter significant challenges. The prevailing approach involves employing a super-resolution reconstruction method to enhance the image, which is then fed into a standard OCR system along with an approximation of the original high-resolution image. However, the researchers propose an end-to-end method that eliminates the need for the super-resolution phase, leading to superior OCR results. Their approach utilizes neural networks for OCR and draws inspiration from the human visual system. Remarkably, their experiments demonstrate that OCR can be successfully performed on scanned images of English text with a resolution as low as 60 dpi, which is considerably lower than the current state of the art. The results showcase an impressive CLA of 99.7\% and a Word Level Accuracy (WLA) of 98.9\% across a corpus comprising approximately 1000 pages of 60 dpi text in diverse fonts. When considering 75 dpi images as an example, the mean CLA and WLA achieved were 99.9\% and 99.4\%, respectively.

\subsection{Tamizhi}
Based on \newcite{munivel2022optical}, digitizing documents from ancient history typically involves OCR. However, OCR for Tamizhi documents poses significant challenges due to the inherent similarities in shape and structure among many characters, along with their subtle variations. The Tamizhi script, also known as Tamil-Brahmi, serves as the precursor to numerous modern Indian scripts and is recognized as one of the oldest scripts in India. Developing an OCR system for Tamizhi script is exceptionally difficult due to the abundance of combined characters, where a character can consist of a single vowel, consonant, or a combination of both. In their research paper, the authors discuss their efforts in creating an OCR system specifically designed for printed Tamizhi documents. The system aims to perform effectively despite various factors, including the poor quality of the documents, the presence of noise, and the diverse formats of the input data. The authors report that their Tamizhi OCR achieves an accuracy rate of 91.12 percent for printed text, demonstrating promising results in recognizing Tamizhi characters.

To summarize, we can mention that up to the time we publish this research, the literature does not report on any efforts made to specifically develop OCR for historical Kurdish documents. Also currently no accessible dataset is available to train OCR systems that are specifically designed to extract text from historical Kurdish documents. That significantly restricts our options when it comes to selecting the most suitable approach for our study. 

To develop an OCR system specifically tailored for historical documents, researchers employed different techniques and strategies such as SVM, LSTM, and CNN. The variability in the obtained results, which reached a maximum of 99.7\% CLA, can be attributed to several contributing factors. These factors include the quality of the dataset used, the specific methodology employed during the development of the OCR system, and the intrinsic complexity of the documents being processed.

The studies that were reviewed in this chapter employed both proprietary datasets that were created by researchers themselves and publicly available datasets. These datasets include TWDB, HWDB, GT4HistOCR, Stockholm Archive, Dunhuang data, Tripitaka, TKH, MTH, and Kana-PRMU. According to the literature in this field, there are ongoing efforts to improve OCR techniques for different kinds of historical documents. 

Based on our research, we identified that LSTM is a widely adopted approach for developing OCR systems with acceptable accuracy.  As a result, we used the latest version of Tesseract, which integrates LSTM functionality, to ensure optimal performance in our project research. Additionally, we discovered the availability of pre-trained models that can be used for fine tuning on our dataset.  Recognizing the similarities between the Kurdish and Arabic scripts, we made the decision to use an Arabic pre-trained model as our base model.

\section{Method}
\label{Sec-Method}
This chapter provides the method of conducting this research. It explains data collection and preparation, the experimental environment and its configurations, and the assessment and evaluation of the outcomes. 

\subsection{Data Collection}
We collect data from different public and private libraries with historical documents. We focus on items published in the early and mid-1900s because the first printing press found in Iraq dating back to the 1920s in Sulaymaniyah by Mandate authorities. It was an old hand-operated letterpress called Chapkhanay Hukumat (Government Press) \cite{hassanpour1992nationalism}. We convert the documents to digital copies. Converting historical documents into digital copies has many issues, and one of them is physical issues. The physical issue with the process involves difficulties from aging, document degradation, and imperfect production processes. Stains, tears, and irregular accumulation of dirt, in addition to artifacts, are some other issues \cite{antonacopoulos2004lifecycle}.

\subsection{Data Preparation}
For optimal performance, Tesseract is best suited for images with a resolution of at least 300 dpi. Therefore, resizing images to meet this requirement can be beneficial. It is worth noting that earlier versions of Tesseract (3.05 and earlier) can handle inverted images, where the background is dark and the text is light, without encountering problems. However, in version 4.x, it is recommended to use images with a light background and dark text for improved performance \cite{googleimagequality}.

\subsection{Preprocessing}
Before conducting OCR, Tesseract incorporates various image processing operations using the Leptonica library. Leptonica is a freely available open-source library encompassing software suitable for various image processing and analysis applications. In most cases, the built-in image processing functionalities of Tesseract effectively prepare the image for OCR. However, there may be instances where additional refinement is necessary, potentially leading to decreased precision. To observe the image processing steps performed by Tesseract, users can enable the configuration variable tessedit\_write\_images and review the processed image. If the resultant image appears to be of low quality, it is possible to apply additional image processing operations before feeding it into Tesseract for improved results \cite{googleimagequality}. 

\begin{itemize}
	\item Inverting images: While previous versions of Tesseract (3.05 and earlier) can handle inverted images (with a dark background and light text) without problems, version 4.x should use a dark background and dark text.
	\item Rescaling: To optimize Tesseract's performance, resizing images to a minimum DPI of 300 is recommended.
	\item Binarization: This process converts an image to black and white. Tesseract internally performs binarization using the Otsu algorithm, but the result may need to be improved, especially if the page background has uneven darkness. Tesseract 5.0.0 introduced Adaptive Otsu and Sauvola, two new Leptonica-based binarization methods.
	\item Noise Removal: Noise refers to unpredictable variations in an image's brightness or color that can hinder text recognition. Tesseract cannot eliminate some forms of noise during binarization, which can lead to decreased accuracy rates.
	\item Dilation and Erosion: Characters with bold or thin features, especially those with serifs, may impact detail recognition and reduce accuracy. Dilation and Erosion operations can be applied to expand or contract the margins of characters against a common background. Erosion can compensate for heavy ink leakage in historical documents and restore characters to their original glyph structure.
	\item Skew Correction: Skewed images can negatively affect Tesseract's line segmentation and OCR quality. Rotating the image to align the text lines horizontally can rectify this issue.
	\item Borders: 
	\begin{itemize}
		\item Missing borders: OCR without a border can cause problems. Adding a minor border (e.g., 10pt) using tools like ImageMagick can help alleviate this issue.
		\item Large borders: Large borders, especially with a single letter/digit or a word on a significant background, can lead to problems ("empty page"). It is recommended to crop the image to fit within the text area with a border of at least 10 points.
		\item Scanning border Removal: Scanned documents often have dark borders, which can be mistakenly interpreted as extra characters, especially if they vary in size, shape, and color.
	\end{itemize}
	\item Transparency / Alpha channel: Certain image formats, like PNG, can incorporate an alpha channel to achieve transparency. Tesseract 4.00, utilizing the Leptonica function pixRemoveAlpha(), can remove the alpha channel by merging the alpha component with a white background. However, this process may lead to issues in specific scenarios, such as performing OCR on movie subtitles. To solve such problems, users might be required to manually eliminate the alpha channel or perform image preprocessing by inverting the colors.
\end{itemize}

\subsubsection{Data Preparation for Tesseract}

Data preparation for Tesseract can be done in two ways: generating the dataset artificially from text files or manually preparing the dataset from image lines. We follow the latter approach. For the images, they should be in TIFF format with the ".tif" extension or PNG format with the extensions ".png", ".bin.png", or ".nrm.png". The transcription need plain text files containing a single line of text. They should have the same name as the corresponding line image but with the extension ".gt.txt" added to the image extension.

\subsection{Environment Setup}
At present, the training process supports Linux as the operating system. While having a multi-core system with OpenMP and Intel Intrinsics support for SSE/AVX extensions is beneficial, but not mandatory. Four cores are considered optimal, but the training can still run on devices with sufficient RAM, albeit slower. The training process does not require a GPU. Apart from the RAM needed for the operating system, having at least 1 GB of additional RAM is recommended. Memory usage can be regulated using the "--max\_image\_MB" command-line option \cite{google5training}.

\subsection{Dataset Preparation}

We choose various schemes for data splitting based on the data that we collect. For the training and evaluation we follow the method that \newcite{idrees2020improve} suggested but we apply it to Tesseract version 5.

\subsection{Evaluation}
Similar to the approach we take for dataset preparation, we follow the method that \newcite{idrees2020improve} suggested for the evaluation as well.

\section{Experiments, Results, and Discussion}
\label{sec-erd}
Initially, we collected some historical publications from the Zaytoon Public Library in Erbil. However, due to the fragile condition of the documents, it was not easy to transfer them into digital format. Then, via the internet, we found the Zheen Center for Documentation and Research in Sulaymaniyahn \url{https://zheen.org}, a facility specializing in scanning and archiving historical documents using unique technologies explicitly designed for that function. After visiting them and explaining our project, they agreed to provide us with digital copies of the earliest Kurdish publications they had in their collection.

\begin{figure}[!h]
	\centering
	\fbox{\includegraphics[width=8cm]{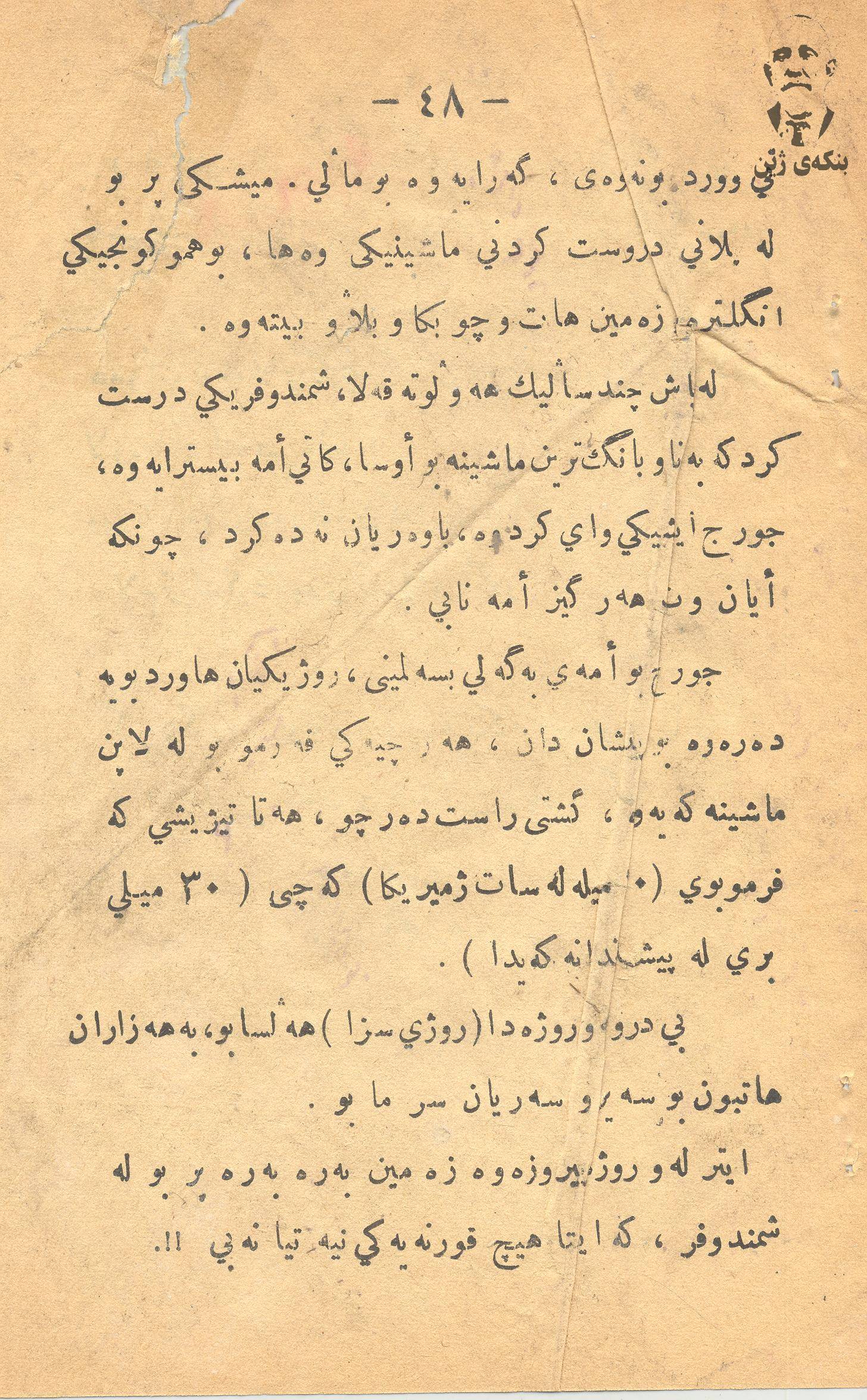}}    
	\caption[Sample page in the book titled 'Awat' published in 1938]{\label{fig:sample_page_in_boo07} Sample page in the book titled 'Awat' published in 1938 (Zheen Center for Documentation and Research)}
\end{figure}

\subsection{Processed Data}
To handle image processing tasks, we utilized a dedicated batch processing tool that was freely available. With this tool, we loaded the images and applied a de-skewing process to correct any skew present in the images. We also performed automatic cropping and converted the images to binary format, saving them in the specified destination directory.

\subsection{Dataset}
After receiving the historical documents from Zheen Center for Documentation and Research in a digital format, we converted the pages into single-line images with respected transcription for the line. We used an Image Processing application to crop lines and saved them in TIFF format.

After converting the pages into image lines (See Figure ~\ref{fig:dataset_image}), we created transcription files for each image line using a text editing program by manually typing what is written in the images. We named the transcription files the same name as the image line with (.gt.txt) postfix (See Figure ~\ref{fig:dataset_transcript}).

\begin{figure}[h]
	\centering
	\frame{\includegraphics[width=\linewidth]{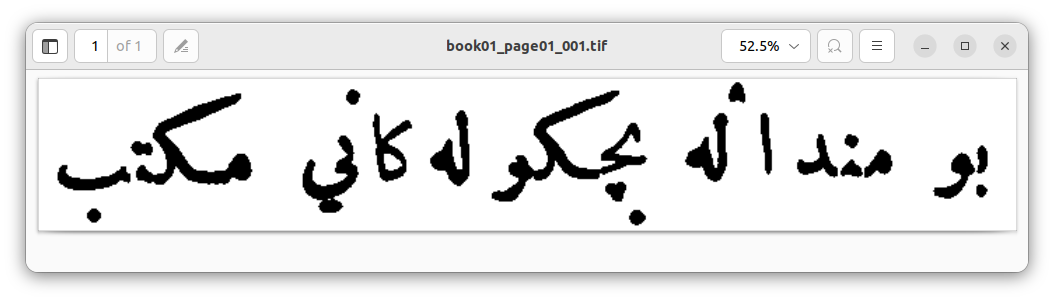}}
	\caption{\label{fig:dataset_image} Image of the extracted line}
\end{figure}

\begin{figure}[h]
	\centering
	\frame{\includegraphics[width=\linewidth]{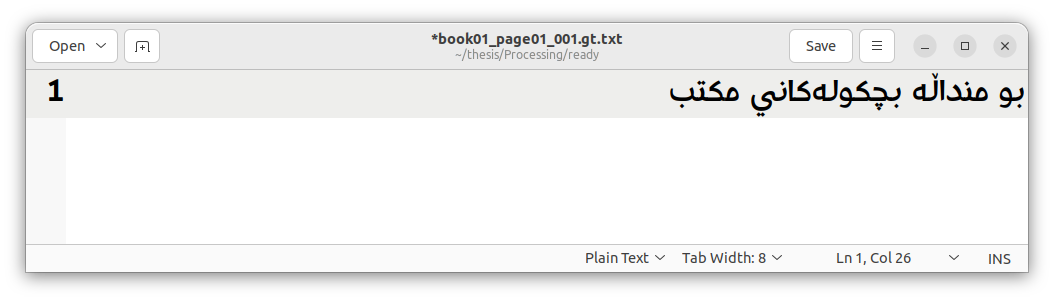}}
	\caption{\label{fig:dataset_transcript} Transcript of the extracted line}
\end{figure}

This way, the dataset for training Tesseract was created, which resulted in 1233 files. Half are the image lines, and the other is the transcription files (See Table ~\ref{tab:collected-data}).

\begin{table}[h]
	\begin{center}
		\caption{Summary of the dataset}
		\label{tab:collected-data}
		\begin{tabular}{|p{7.0cm}|p{1.3cm}|p{2.2cm}|}
			\hline
			{Publication} & {Year} & {No of Lines} \\ \hline 
           {\small{\RL{دەستە گوڵی لاوانە}}} (Deste Gull\^i Lawane) & 1939 & 273 \\ \hline        
			{\small{\RL{کتاب سەما و زەمین}}} (Kitab Sema \^u Zem\^in)  & 1936 & 227 \\ \hline    
			{\small{\RL{ئاوات}}} (Awat) & 1938 & 355 \\ \hline    
			{\small{\RL{عیلمی ژمارە}}} ('lm\^i Jmare) & 1936 & 129 \\ \hline    
			{\small{\RL{خوێندنەوەی کوردی بۆ پۆلی سێیەم}}} (Xw\^endineweyi Kurd\^i Bo Pol\^i S\^eyem) & 1947 & 219 \\ \hline    
			\multicolumn{2}{|l|}{Total} & {1233} \\ \hline    
		\end{tabular}
	\end{center}
\end{table}

\subsection{Experiments}
In this section, we provide details of the steps taken to prepare our environment, the training process of the model, and other relevant aspects.

\subsubsection{Environment Setup}
For this training environment, we used Ubuntu 22.04.2 LTS (Jammy Jellyfish). We cloned the \textit{tesstrain} from \url{https://github.com/tesseract-ocr/tesstrain} and we trained the model using our prepared dataset.

\subsection{Results and Evaluation}
After completing the training, we evaluated the model using different methods. In this section, we show the results for each evaluation method. During the training process, the trainer produces a report that outlines the model's accuracy every 100 iterations. Once the training was completed, we assessed the model using Tesseract evaluation and obtained a minimal training error rate (BCER) value of 0.755\%.

We randomly chose a subset of pages from the collected data which were not utilized in the model's training and testing. These pages were manually transcribed, and the page images and their corresponding manual transcriptions were submitted to Ocreval for evaluation. The outcomes of this evaluation can be observed in Figures ~\ref{fig:book01_page14}, ~\ref{fig:book01_page14_gt}, ~\ref{fig:book01_page14_actual}, and ~\ref{fig:ocreval_01}.

\begin{figure}
	\centering\frame{\includegraphics[width=15cm]{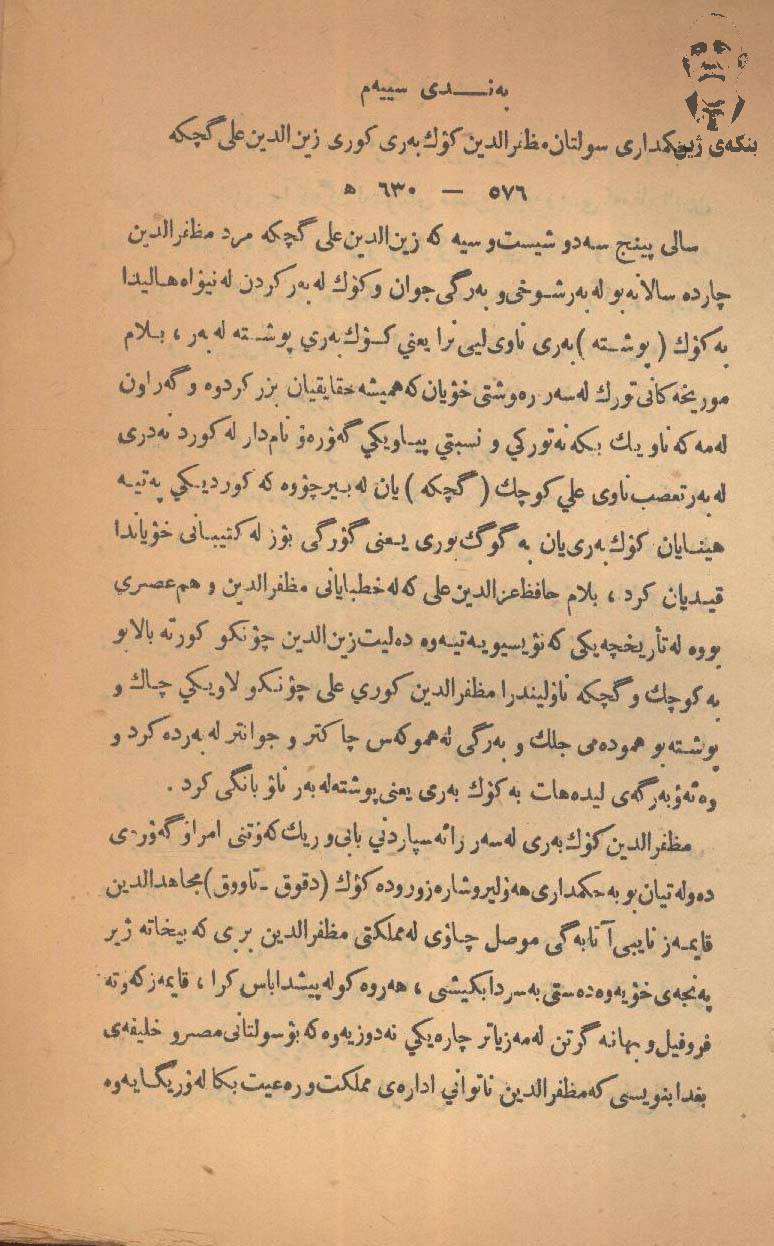}}
	\caption[A sample page from the book titled 'Awreky Pashawa' published in 1930]{\label{fig:book01_page14} A sample page from the book titled 'Awreky Pashawa' published in 1930 (Zheen Center for Documentation and Research)}
\end{figure}

\begin{figure}
	\centering\noindent\frame{\includegraphics[width=\linewidth]{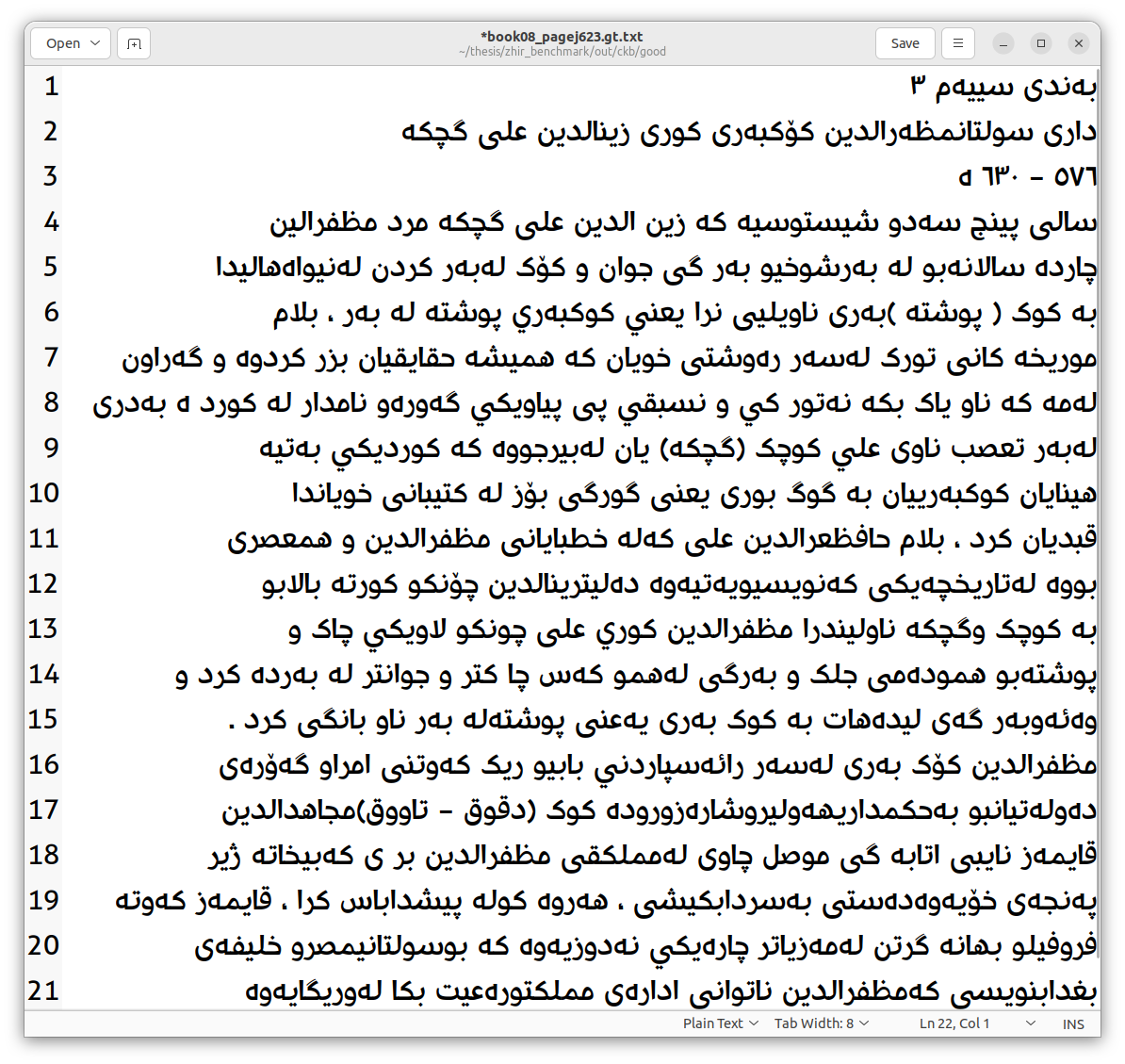}}
	\caption{\label{fig:book01_page14_gt} Manual transcription of the page}
\end{figure}

\begin{figure}
	\centering\noindent\frame{\includegraphics[width=\linewidth]{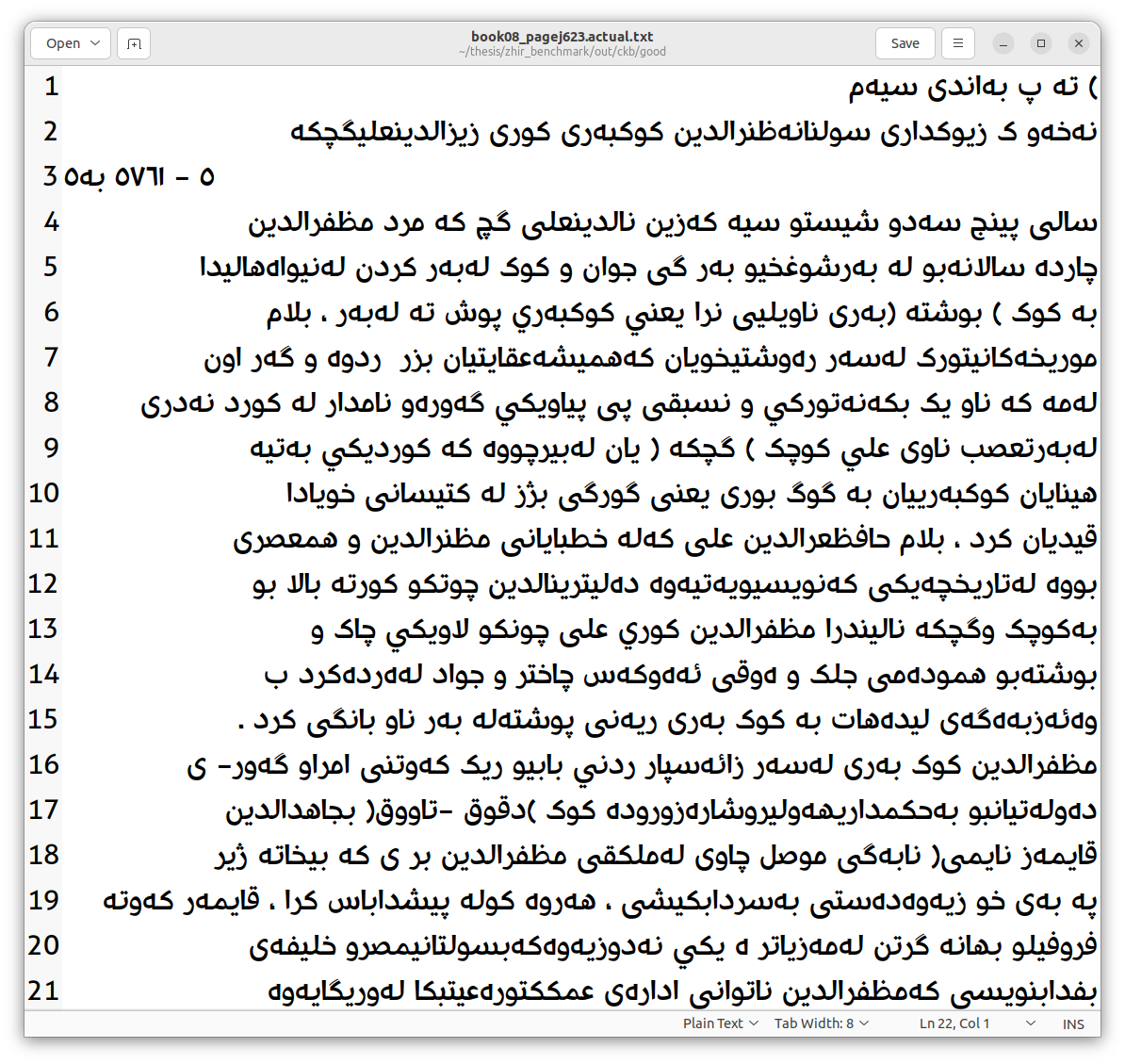}}
	\caption{\label{fig:book01_page14_actual} The transcription generated by our model}
\end{figure}

\begin{figure}[h]
	\centering\frame{\includegraphics[width=\linewidth]{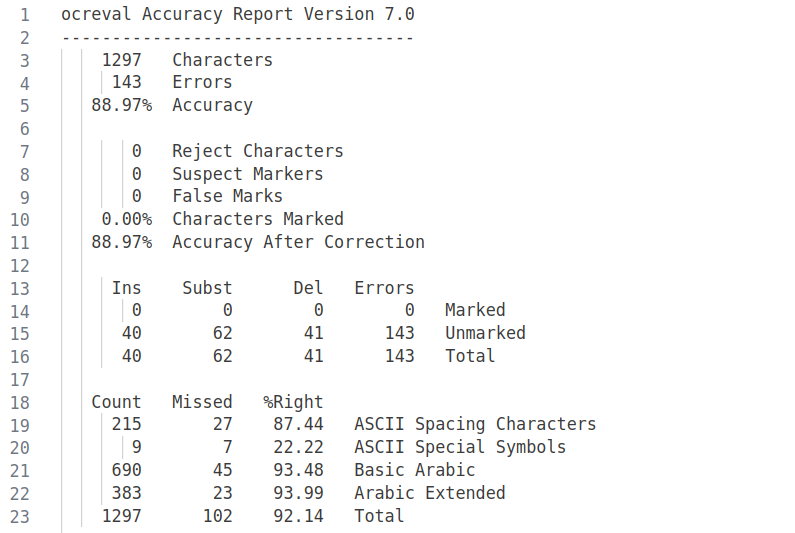}}
	\caption{\label{fig:ocreval_01} Result of Ocreval tool}
\end{figure}

\begin{table}[h]
	\begin{center}
		\caption{Ocreval result}
		\label{fig:ocreval_result_table}
		\begin{tabular}{|p{7.0cm}|p{1.5cm}|p{2.0cm}|p{2.0cm}|p{1.9cm}|}				
			\hline
			{Publication} & {Year} & {Chars} & {Errors} & {Accuracy}  \\ \hline 
			{\small{\RL{دەستە گوڵی لاوانە}}} (Deste Gull\^i Lawane) & 1939 & 667 & 105 & 84.26 \% \\ \hline 
			{\small{\RL{محاسەبەی نیابەت}}} (Mihasbeyi Niyabit) & 1928 & 787 & 152 & 80.69\% \\ \hline 
			{\small{\RL{ئاوات}}} (Awat) & 1938 & 735 & 157 & 78.64\% \\ \hline 
			{\small{\RL{ئاورێکی پاشەوە}}} (Awrr\^ek\^i Pa\c{s}ewe) & 1930 & 1297 & 143 & 88.97\% \\ \hline 
			\multicolumn{2}{|l|}{Total} & {3486} & {557}  & {84.02\%}  \\ \hline 
		\end{tabular}
	\end{center}
\end{table}	

\subsection{Discussion}
The limited availability of resources presented significant challenges during our data collection process. Converting the collected data into a digital format proved to be an additional obstacle, for which we received support from the Zheen Center for Documentation and Re- search. Manual transcription of the documents posed considerable difficulty due to unclear text, non-standard spacing between words and characters, and unique vocabulary influenced by Arabic letters and terminologies. Also, we discovered that the system has challenges in properly extracting text from multi-column pages and mathematical equations.

We retrained an existing Arabic model using our unique Kurdish dataset in this research, which yielded remarkable outcomes. Considering our findings, if we further train the model on a larger dataset, it has the potential to produce results suitable for production use. Such a model can significantly aid libraries and centers in effectively extracting text from historical documents.

\section{Conclusion}
\label{Sec-Conc}
The primary motivation for this study stems from the significant amounts of historical documents stored in libraries that still need to be processed. The lack of processing capabilities has led to exploring OCR technology for Kurdish, a low-resource language. Implementing OCR for extracting text from historical documents in Kurdish would greatly enhance available resources.

Extensive research was conducted to assess existing OCR systems for Kurdish and other languages worldwide. The investigation focused on previous work, accuracy, and underlying technology. It was determined that Tesseract was a suitable option for this research.

Once the technology was identified, efforts were made to collect digital copies of historical documents printed before 1950. This task proved challenging, as locating documents and converting them into digital format presented additional hurdles. Fortunately, the Zheen Center for Documentation and Research in Sulaymaniyah, which specializes in archiving historical documents, provided some books in the form of digital copies.

Upon receiving the digitized copies, a dataset was created to train the Tesseract model. Text lines were extracted from the pages, transcribed individually, and subjected to preprocessing to prepare the dataset.

With a dataset of 1233 lines, the model was trained based on the Arabic model. Following the training, the model's performance was evaluated using various methods. Tesseract's built-in evaluator lstmeval indicated a CER of 0.755\%. Additionally, Ocreval demonstrated an average character accuracy of 84.02\%. Finally, an in-house web application was developed to provide an easy-to-use interface for end-users, allowing them to interact with the model by inputting an image of a page and extracting the text.

This model could be a valuable tool for libraries and centers, enabling them to extract text from historical documents and perform further processing effectively.

\subsection{Challenges and Limitations}

Following is the list of main challenges and limitations we faced during this research:

\begin{itemize}
	\item The limited availability of resources posed significant challenges during our data collection process. Converting the collected data into a digital format proved an additional obstacle. Manual transcription of the documents was difficult due to unclear text, non-standard spacing, and unique vocabulary influenced by Arabic letters and terminologies. We attempted to create the dataset synthetically, crafting a small tool that assembled letters from a given collection of character images. Regrettably, the outcomes were unsatisfactory, and given our time constraints, we discontinued this approach.
	
	\item The non-standard spacing between the words and characters was challenging for transcribing the documents and needed to be more apparent for the model. The model interpreted the excessive gaps between characters or words as space characters. In contrast, in other cases where there should have been a space character, the minimal spacing went unnoticed by the model.
	
	\item Extracting text from multi-column pages was another limitation of the model.
	
	\item Recognizing mathematical equations was another limitation of our model.
\end{itemize}

Considering the challenges and limitations and based on the discussion on the results, we are interested in exploring several areas in the future as follows: 

\begin{itemize}
	\item Expanding the dataset is an aspect that requires further attention and effort.
	\item An observed issue pertained to the misalignment of spaces between words and characters. To address this, a post-processing phase is suggested for rectifying the misaligned space characters.
	\item Ocr'ing the multi-column pages property is another area requiring more effort.
	\item Extracting mathematical equations accurately.
\end{itemize}

\section*{Online Resources}

The dataset is partially publicly available for non-commercial use under the CC BY-NC-SA 4.0 license at \url{https://github.com/KurdishBLARK/OCR4OldTextsInSorani}.

\section*{Acknowledgments}
We would like to extend our gratitude to the Zheen Center for Documentation and Research in Sulaymaniyah, Kurdistan Region, Iraq for their generous support in providing us with digital copies of certain historical publications.

\bibliographystyle{lrec}
\bibliography{KurdishOldTextsOCR}

\end{document}